\documentclass{article}

\usepackage[final]{arxiv_2021}

\usepackage[utf8]{inputenc} 
\usepackage[T1]{fontenc}    
\usepackage{hyperref}       
\usepackage{url}            
\usepackage{booktabs}       
\usepackage{amsfonts}       
\usepackage{nicefrac}       
\usepackage{microtype}      
\usepackage{xcolor}         

\usepackage{booktabs} 
\usepackage{enumitem}
\usepackage{amsmath}        
\usepackage{amsthm}         
\usepackage{amssymb}        
\usepackage{mathtools}      
\usepackage[ruled,vlined,linesnumbered]{algorithm2e}    
\usepackage{algpseudocode}    
\usepackage{graphicx}       
\usepackage{subcaption}     
\usepackage{appendix}       
\usepackage{float}          

\usepackage{multirow}       
\usepackage{lipsum}
\usepackage{xcolor}
\usepackage{wrapfig}
\usepackage{lscape}

\newcommand\extrafootertext[1]{%
    \bgroup
    \renewcommand\thefootnote{\fnsymbol{footnote}}%
    \renewcommand\thempfootnote{\fnsymbol{mpfootnote}}%
    \footnotetext[0]{#1}%
    \egroup
}

\newcommand{\model}{\textsc{ProGen2}\xspace}
\newcommand{\jf}{\textsc{JAXformer}\xspace}

\title{ProGen2: Exploring the Boundaries of Protein Language Models}

\author{%
  Erik Nijkamp*\\
  Salesforce Research 
   \And
   Jeffrey Ruffolo* \\
   Johns Hopkins University  
   \And
   Eli N. Weinstein \\
   Columbia University  
   \And
   Nikhil Naik \\
   Salesforce Research
   \And
   Ali Madani \\
   Salesforce Research 
}

\begin{document}

\maketitle

\begin{abstract}
Attention-based models trained on protein sequences have demonstrated incredible success at classification and generation tasks relevant for artificial intelligence-driven protein design.
However, we lack a sufficient understanding of how very large-scale models and data play a role in effective protein model development.
We introduce a suite of protein language models, named ProGen2, that are scaled up to 6.4B parameters and trained on different sequence datasets drawn from over a billion proteins from genomic, metagenomic, and immune repertoire databases.
ProGen2 models show state-of-the-art performance in capturing the distribution of observed evolutionary sequences, generating novel viable sequences, and predicting protein fitness without additional finetuning.
As large model sizes and raw numbers of protein sequences continue to become more widely accessible, our results suggest that a growing emphasis needs to be placed on the data distribution provided to a protein sequence model.
We release the ProGen2 models and code at \href{https://github.com/salesforce/progen}{https://github.com/salesforce/progen}.
\extrafootertext{*Equal Contribution \\ Correspondence to ali@madani.ai}
\end{abstract}

\section{Introduction}
Proteins are the workhorse of life -- performing essential and versatile functions critical to sustain human health and the environment. Engineering proteins for our desired purposes enables use-cases in industries across pharmaceuticals, agriculture, specialty chemicals, and fuel. Current tools for protein engineering are limited and, as a consequence, mainly rely on directed evolution~\citep{arnold1998design}, a process of stochastically mutating a starting/wild-type sequence, measuring each variant, and iterating until sufficiently optimized for improved function, also referred to as fitness.

Nature as an underlying generative process has yielded a rich, complex distribution of proteins. Due to exponentially-broken barriers in DNA sequencing, we now collect natural sequences at a previously-unimaginable pace. In parallel, we have seen machine learning models perform exceedingly well at capturing data distributions of images and natural language \citep{saharia2022photorealistic,brown2020language}. In particular, the transformer~\citep{vaswani2017attention} has proven to be a powerful language model and can serve as a universal computation engine~\citep{lu2021pretrained} across data modalities.

Our research moonshot is to build machine learning models that effectively learn from nature and its underlying principles for functional protein engineering and design. Protein language models have shown promise in representation learning for classification, regression, and generation purposes~\citep{rives2021biological, rao2021msa,rao2019evaluating,madani2020progen,meier2021language,elnaggar2020prottrans,brandes2022proteinbert}. Along this pursuit, we perform a study on the effect of very large-scale models and data. In short:
\begin{itemize}
    \item We train a suite of models ranging from 151M to 6.4B parameters (the largest published for a single protein transformer) on different train sets collectively totaling 1B protein sequences from genomic, metagenomic, and immune repertoire databases. 
    \item We analyze the generations from universal and family-specific models through predicted structural and biophysical properties.
    \item We examine fitness prediction on existing experimental datasets which motivate hypotheses on the role of data distribution and alignment in protein language modeling. 
\end{itemize}

\section{Related Work}
\paragraph{Large-scale language modeling:}

Language modeling tries to capture the notion that some sequences are more likely than others by density estimation. Since language has a natural total order over symbols, the joint distribution over symbols can be modeled as the product of conditionals, which allows for tractable sampling and likelihood estimation. For such causal language models, transformer models equipped with self-attention mechanisms~\citep{bahdanau2014neural} have shown to be particularly well suited to capture dependency among sequence elements while being capable to scale vast amounts of model parameters~\citep{kaplan2020scaling,hoffmann2022training}.

Remarkably, large language models (LLM) with a vastly increased number of model parameters revealed two intriguing phenomena.

First, scaling laws~\citep{kaplan2020scaling,hoffmann2022training} dictate the cross-entropy of LLMs as simple power-law relations between the amount of compute, observations, and model parameters, thus allowing to calculate a compute-optimal configuration and guaranteeing a certain perplexity as a proxy for performance on down-stream tasks.

Second, few-shot learning~\citep{brown2020language} models tasks as auto-regressive sampling conditional on a small set of examples (or shots). Notably, LLMs possess the capacity to solve the intended task by increasing the number of parameters without task-specific fine-tuning of the model parameters. These few-shot abilities appear to emerge under certain parameter thresholds~\citep{wei2022emergent}, which motivates the exploration of such capabilities for the domain of protein engineering.

In this work, we adopt causal language models in the form of auto-regressive decoders for the modeling of proteins. The raw amino acid sequences which constitute a protein are considered as observed sequences for the maximum likelihood-based learning. The problem of conditional protein generation is naturally cast as a next-token prediction task.

\paragraph{Protein sequence generation:} 
Methods for generating protein sequences that are functional and have desired properties have recently seen tremendous progress. Simple, traditional methods that leverage multiple sequence alignments of similar proteins, such as ancestral sequence reconstruction \citep{gumulya2018engineering}, have demonstrated the ability to generate useful proteins but are limited in scope. A host of statistical and machine learning techniques exist to access a larger sequence space. Most still train on a fixed protein family to capture coevolutionary signals present within a set of homologous sequences -- ranging from direct coupling analysis techniques \citep{russ2020evolution} to generative adversarial networks \citep{repecka2021expanding}. More versatile models trained on unaligned and unrelated sequences have emerged \citep{shin2021protein} for functional sequence design.

Language models, in particular, provide a powerful architecture to learn from large sets of amino acid sequences across families for the purpose of generating diverse, realistic proteins \citep{madani2020progen, ferruz2022deep}. Sequences generated by protein language models (PLMs) are typically predicted to adopt well-folded structures, despite diverging specifically in sequence space. 
PLMs can be further focused on specific families of interest by finetuning on a subset of relevant proteins. In prior work, finetuning the ProGen model on a set of lysozyme families yielded proteins retaining functional behavior, and even rivaling that of a natural hen egg white lysozyme \citep{madani2021deep}.
Similar strategies have been employed for domain-specific PLMs, such as the antibody-specific IgLM model \citep{shuai2021generative}. By conditioning on chain type and species-of-origin, IgLM is capable of generating diverse sets of antibodies resembling those of natural immune repertoires.

Recent work has also focused on the generation of amino acid sequences conditioned on a fixed protein backbone structure. Structure-conditioned sequence generation is typically approached as an encoding of backbone structure followed by auto-regressive decoding of amino acids. Early work represented the protein backbone as a proximity graph, which was encoded and decoded using a graph transformer model \citep{ingraham2019generative}. This approach improved sequence recovery (the fraction of generated residues matching the original protein sequence) over the popular Rosetta protein design software \citep{leaver2011rosetta3}, while reducing compute time by an order of magnitude. Motivated by this work, ESM-IF1 \citep{hsu2022learning} replaced the encoder network with an rotation-equivariant model \citep{jing2020learning} and scaled the network to 142M parameters, observing further improvements to sequence recovery with increased model capacity. Finally, ProteinMPNN \citep{dauparas2022robust} utilized a rotation-invariant message passing network for encoding structures, followed by a novel order-agnostic decoding of sequences. Extensive experimental validation demonstrated that sequences generated by ProteinMPNN dramatically improved soluble expression and folded into their intended structures \citep{wicky2022hallucinating}.

\begin{table}[t]
\begin{center}
\begin{small}
\noindent\makebox[\textwidth]{%
\begin{tabular}{lrrrrr}
\toprule
& \multicolumn{5}{c}{Model}\\\cmidrule{2-6}
Hyper-parameter & \model-small & \model-medium & \model-base & \model-large & \model-xlarge\\
\midrule
Number of params & 151M & 764M & 764M & 2.7B & 6.4B\\ 
Number of layers & 12 & 27 & 27 & 32 & 32 \\
Number of heads & 16 & 16 & 16 & 32 & 16 \\
Head dimensions & 64 & 96 & 96 & 80 & 256 \\
Context length & 1,024 & 1,024 & 2,048 & 1,024 & 1,024 \\
\midrule
Batch size & 500k & 500k & 500k & 500k & 1M \\
Learning rate & 6.0e-4 & 2.5e-4 & 2.0e-4 & 0.8e-4 & 0.1e-4 \\
Weight decay & 0.1 & 0.1 & 0.1 & 0.1 & 0.1 \\
Grad norm clip & 1.0 & 1.0 & 0.8 & 0.8 & 0.8 \\
Warm-up steps & 3,000 & 3,000 & 10,000 & 10,000 & 10,000 \\
Total steps & 350,000 & 350,000 & 400,000 & 400,000 & 350,000 \\
\bottomrule
\end{tabular}
}
\end{small}
\end{center}
\vspace{5mm}
\caption{Choice of hyper-parameters for model specification and optimization for the family of \model causal language models for protein engineering.}
\label{table:training}
\end{table}

\paragraph{Protein fitness prediction:} 
Understanding the functional effects of sequence mutations is critical for the rational design of proteins. Such effects can be quantified by deep mutational scanning (DMS) experiments, which systematically evaluate the impact of mutations on a given measured attribute \citep{fowler2014deep}. However, DMS experiments typically require characterization of many thousands of mutations, making them costly and time-consuming endeavors, which are limited in terms of the full combinatorial space of possible useful mutations. Machine learning models trained on protein sequences offer an alternative means of capturing the functional, i.e. fitness, landscape of protein sequences. These methods typically fit into one of two categories: family-specific models trained on aligned sequences or universal models trained on unaligned sequences.

Covariation of residues within multiple sequence alignments (MSAs) of related proteins provide insights into the functional constraints on the sequences. This insight has been leveraged to train family-specific models for prediction of protein fitness. EVMutation \citep{hopf2017mutation} trained a Potts model (a second-order approach capturing covariation between pairs of residues) to predict fitness of variants. However, protein fitness is often determined by higher-order dependencies than can be described by pairwise relationships. DeepSequence \citep{riesselman2018deep} applied a variational autoencoder (VAE) to MSAs, enabling the model to distill such higher-order interactions into a latent representation, which was used for classification according to the evidence lower bound (ELBO). Extending this approach with a Bayesian VAE, EVE \citep{frazer2021disease} achieved further improvements to variant effect prediction.

Models based on alignments of sequence face several key challenges limiting their application to protein engineering tasks. First, for proteins with few evolutionary neighbors, the MSA is likely to be shallow and contain little information about functional constraints. Second, for some families of proteins (such as antibodies), there are many sequences available, but they are non-trivial to align. Finally, evaluation of novel variants requires that new sequences be aligned to the MSA used for training; this can be challenging in cases with significant insertions or deletions (indels). These limitations prompted the development of fitness predictors based unaligned sets of sequences, particularly transformer models trained on large databases of protein sequences. ESM-1v \citep{meier2021language} tasks a transformer encoder model trained via masked-language modeling with estimating heuristic likelihood of mutations relative to the wild type sequences. In a zero-shot setting, ESM-1v approached the performance of family-specific models for DMS variant prediction, and outperformed such models with finetuning. Autoregressive PLMs have also been applied to fitness prediction \citep{shin2021protein}. These models are intrinsically capable of modeling indels, as well as epistatic mutations. The RITA family of models \citep{hesslow2022rita} demonstrated that not only do autoregressive PLMs effectively estimate protein fitness, but performance also be further improved by scaling model capacity. Tranception \citep{notin2022tranception} demonstrated that combining autoregressive language models with retrieval \citep{borgeaud2021improving} capabilities provides a means of enhancing a generalist model with family-specific information from MSAs at inference. With retrieval, Tranception models outperform family-specific models for protein fitness prediction, particularly for shallow MSAs.

\begin{table}[t]
\centering
\begin{tabular}{l|c|c|c}
\toprule
Model Name     & Parameters (N) & Test-max90 (ppl) & Test-max50 (ppl) \\
\midrule
\model-small  & 151M           & 12.9          & 15.0           \\
\model-medium & 764M           & 11.2          & 14.3           \\
\model-large  & 2.7B           & 11.1          & 14.4           \\
\model-xlarge & 6.4B           & 9.9           & 13.9           \\
\bottomrule
\end{tabular}
\vspace{5mm}
\caption{Increasing number of parameters allows the model to better capture the distribution of observed evolutionary sequences. Performance is measured as the perplexity of held-out test sequences at various maximum sequence identity thresholds, i.e. test-max50 is more difficult and out-of-distribution.}
\label{tab:testperformance}
\end{table}

\section{Methods}
\subsection{Model}

Our models are autoregressive transformers with next-token prediction language modeling as the learning objective. The family of \model models is trained in various sizes with 151M, 764M, 2.7B, and 6.4B parameters.

The architecture follows a standard transformer decoder with left-to-right causal masking. For the positional encoding, we adopt rotary positional encodings~\citep{su2021roformer}. For the forward pass, we execute the self-attention and feed-forward circuits in parallel for improved communication overhead following~\citep{gpt-j}, that is, $x_{t+1} = x_t + {\rm mlp}({\rm ln}(x_t + {\rm attn}({\rm ln}(x_t))))$ is altered to $x_{t+1} = x_t + {\rm attn}({\rm ln}(x_t)) + {\rm mlp}({\rm ln}(x_t))$ for which the computation of self-attention, ${\rm attn()}$, and feed-forward, ${\rm mlp()}$, with layer-norm, ${\rm ln()}$, is simultaneous.

Table~\ref{table:training} summarizes the model specifications and choice of hyper-parameters for the optimization such models. The choice of the hyper-parameters was informed by~\cite{brown2020language}, however, the number of layers is reduced with a small number of self-attention heads of relatively high dimensionality to improve overall utilization of the TPU-v3 compute. As explored in~\cite{brown2020language,gpt-j,nijkamp2022conversational}, these variations introduce insignificant degradation of perplexity for sufficiently large models, while significantly improving computational efficiency.

\subsection{Data}
The standard \model models are pretrained on a mixture of Uniref90 \citep{suzek2015uniref} and BFD30 \citep{steinegger2018clustering} databases. Uniref90 are cluster representative sequences from UniprotKB at 90\% sequence identity.  The BFD30 dataset is approximately $1/3$ the size of Uniref90, majority from metagenomic sources, commonly not full-length proteins, and clustered at 30\% sequence identity. For the \model-BFD90 model, Uniref90 is mixed with representative sequences with at least 3 cluster members after clustering UniprotKB, Metaclust, SRC, and MERC at 90\% sequence identity. This BFD90 dataset is approximately twice the size as Uniref90.

To train the antibody-specific \model-OAS, we collected unpaired antibody sequences from the Observed Antibody Space (OAS) database \citep{olsen2022observed}. OAS is a curated collection of 1.5B antibody sequences from eighty immune repertoire sequencing studies, which contains heavy and light chain sequences from six species (humans, mice, rats, camel, rabbit, and rhesus). The sequences in OAS possess a significant degree of redundancy, due both to discrepancies in the sizes of its constituent studies, as well as the innate biological redundancy of antibody sequences within organisms. To reduce this redundancy, we clustered the OAS sequences at 85\% sequence identity using Linclust \citep{steinegger2018clustering}, yielding a set of 554M sequences for model training. Alignment coverage in Linclust was calculated with respect to the target sequence ("cov-mode 1"), with all other parameters set to their default values.

All samples are provided to the model with a 1 or 2 character token concatenated at the N-terminal and C-terminal side of the sequence. Each sequence is then provided as-is and flipped. For a given batch, proteins are concatenated with others to fill the maximum token length during training.

\begin{figure}[t]
    \centering
    \includegraphics[width=1.0\textwidth]{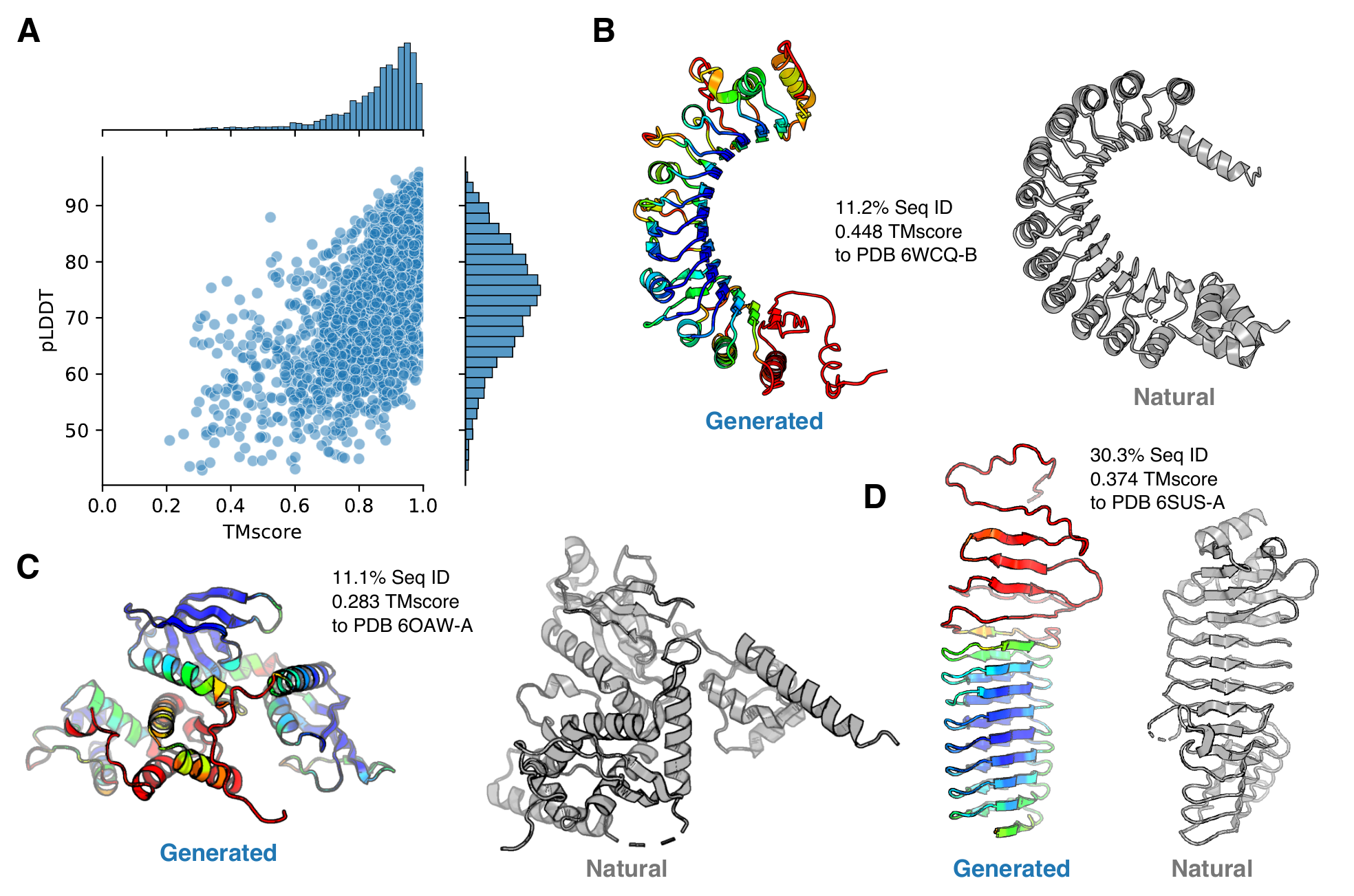}
    \caption{Generating from a pretrained language model trained on a universal protein dataset. (A) Relationship between AlphaFold2 prediction confidence (pLDDT) and similarity to natural protein structures in the PDB (TMscore). (B-D) Comparison of predicted structures for generated sequences (left, colored by pLDDT) and their closest structural counterparts in the PDB (right, gray). Sequence identities and TMscores are calculated against the closest structural matches in the PDB. (B) Solenoid-fold protein generated by the model, with very low sequence identity and high structural similarity to a leucine-rich repeat protein. The generated protein replaces several alpha helices on the outer edge of the fold with beta sheets, resulting in a looser curvature compared to that of its most similar natural counterpart. (C) Multi-domain $\alpha$+$\beta$-fold generated protein with very low sequential or structural similarity to natural proteins. (D) Generated protein resembling RTX domain of adenylate cyclase toxin. The generated protein contains more ordered secondary structure (higher beta sheet content, shorter loops) than other beta-roll folds found in the PDB.}
    \label{fig:pretrain_generations}
\end{figure}

\subsection{Training}


The scaling of large language models requires data and model parallelism. Google's TPU-v3 hardware with a high-speed toroidal mesh interconnect naturally allows for efficient parallelism. To efficiently utilize the hardware, the training of the models is implemented in JAX~\cite{jax2018github}. For parallel evaluation in JAX the $pjit()$\footnote{\url{https://jax.readthedocs.io/en/latest/_modules/jax/experimental/pjit.html}} operator is adopted. The operator enables a paradigm named single-program, multiple-data (SPMD) code, which refers to a parallelism technique where the same computation is run on different input data in parallel on different devices.\footnote{\url{https://jax.readthedocs.io/en/latest/jax-101/06-parallelism.html}} Specifically, $pjit()$ is the API exposed for the XLA SPMD partitioner in JAX, which allows a given function to be evaluated in parallel with equivalent semantics over a logical mesh of compute.

Our library \jf recruits a designated coordinator node to orchestrate the cluster of TPU-VMs with a custom TCP/IP protocol. For data parallelism, the coordinator partitions a batch and distributes the partitions to the individual TPU-VMs. For model parallelism, a partitioning scheme is adopted where parameters are sharded across MXU cores inside a physical TPU-v3 board and replicated across boards following \cite{shoeybi2019megatron, gpt-j}.


For the pre-training of the \model models, Table~\ref{table:training} summarizes the hyper-parameters. We adopt the Adam~\citep{kingma2014adam} optimizer with $(\beta_1, \beta_2, \epsilon) = (0.9, 0.999, 1\mathrm{e}{-08})$ and global gradient norm clipping~\citep{pascanu2013difficulty} of $0.8$ and $1.0$. The learning rate function over time follows GPT-3~\citep{brown2020language} with warm-up steps and cosine annealing.


Notably, the cross-entropy appeared to diverge from the projected power-law relation over time when following standard configurations detailed in~\cite{brown2020language}. In particular, an increasing the global norm of the gradient as an indicator for a divergence from the expected log-log linear behavior of cross-entropy over time was observed. Decreasing the learning rate, increasing weight-decay (or equivalently $\ell_2$-regularization under re-parameteriztation) and decreasing the gradient norm clipping factor resulted in a near-constant global norm of the gradient which stabilized training. 


For the fine-tuning of the \model models, the training is continued from a converged model. The state of the optimizer is re-initialized such Adam's moving averages for the first and second moment estimators are set to zero. The learning rate decay function is adjusted such that initial learning-rate is decreased by a factor of 5. The fine-tuning covers at most two epochs over the fine-tuning dataset to avoid over-fitting.

\subsection{Evaluation}

To investigate the properties of sequences generated by the \model family of models, we sampled complete protein sequences in three settings: universal generation after pretraining, fold-specific generation after finetuning, and antibody generation after pretraining on only antibody sequences. For universal protein generation, we sampled 5,000 sequences from the \model-xlarge model. A diverse set of sequences was sampled using a Cartesian product of temperature ($T\in\{0.2,0.4,0.6,0.8,1.0\}$) and nucleus sampling ($P\in\{0.5,0.7,0.9,1.0\}$) parameters. To understand the effects of architecture-specific finetuning on sequence generation, we compared the sequences produced by the \model-large model after one and two epochs of finetuning. Using a similar strategy as for universal protein generation, 10,000 sequences were generated using a Cartesian product of temperature ($T\in{0.2, 0.4, 0.6, 0.8, 1.0}$) and nucleus sampling ($P\in{0.7, 0.9, 1.0}$) parameters for both model checkpoints. The structures of all generated sequences were predicted with AlphaFold2 \citep{jumper2021highly} with single-sequence inputs (no MSAs) and structural templates from the PDB \citep{berman2000protein}. For universal generations, we ran AlphaFold2 with three recycles while for finetuned generations only one recycle was used. All structures were predicted with the full five-model ensemble (using the pTM models) and the top-ranked structures for each sequence were considered for structural analysis. Similarity of predicted structures to observed proteins in the PDB was measured by calculating the TMscore \citep{zhang2004scoring} using Foldseek \citep{van2022foldseek}. For universal generations, we report the sequence identity against the most structurally similar protein reported by Foldseek. For finetuned generations, we calculated the sequence identity against the finetuning dataset using MMseqs2 \citep{steinegger2017mmseqs2}.

Antibody sequences were generated using the \model-OAS model after pretraining on a set of variable-fragment sequences from the OAS. We evalauted sequences generated by the model with and without initial-residue prompting. A set of 52K unprompted sequences was generated using sampling parameters from a Cartesian product of temperature ($T\in\{0.2,0.4,0.6\}$) and nucleus sampling probability ($P\in\{0.5,0.7,0.9,1.0\}$). An additional 470K full-length sequences were generated by initializing the sequence with a three-residue motif commonly observed in human heavy chain antibody sequences (EVQ). Prompted sequences were similarly generated using a Cartesian product of temperature ($T\in\{0.2,0.4,0.6,0.8,1.0\}$) and nucleus sampling ($P\in\{0.5,0.7,0.9,1.0\}$) parameters. The sequence identity of generated sequences against the training dataset was calculated with MMseqs2 \citep{steinegger2017mmseqs2}. IgFold \citep{ruffolo2022fast} was used to predict structures for all generated antibody sequences. The full four-model ensemble of IgFold models was used for predictions, with PyRosetta \citep{chaudhury2010pyrosetta} refinement applied to model outputs. Aggregation propensities of generated sequences were measured by calculating the SAP score \citep{chennamsetty2010prediction} of the predicted structures. Solubility profiles were calculated based on sequence using the public CamSol-intrinsic \citep{sormanni2015camsol} web server.

Two test sets at differing levels of difficulty were constructed to examine language modeling performance. Test-max90 and Test-max50 correspond to representative sequences from held-out clusters from the Uniref90+BFD30 set of sequences at 90\% and 50\% sequence identity respectively.

To assess zero-shot fitness prediction ability, we evaluate on three sets of experimentally-measured protein landscapes: narrow, wide, and antibody-specific. The narrow landscape set is comprised of the \citet{riesselman2018deep} datasets as provided by the authors of \citet{hesslow2022rita} and generally includes variants that are one or two substitutions away from a given wild-type/natural sequence. The wide landscape set involves larger edit distances and are comprised of the \citet{dallago2021flip} proteins, chorismate mutase proteins from \citet{russ2020evolution}, and the GFP test set proteins from \citet{rao2019evaluating}. 

Lastly, for the antibody-specific landscape, we compiled a dataset consisting of binding, expression, and thermal stability measurements for variants derived from eight distinct antibodies. We collected expression and antigen-binding enrichment measurements for variants of the anti-VEGF g6 antibody from a DMS study \citep{koenig2017mutational}. From a second DMS study, we collected binding enrichment measurements for variants of the d44 anti-lysozyme antibody \citep{warszawski2019optimizing}. Binding affinity ($K_D$) and thermal stability measurements ($T_M$) for the remaining six antibodies (C143, MEDI8852UCA, MEDI8852, REGN10987, S309, and mAb114) were drawn from a recent study on antibody affinity maturation using pretrained language models \citep{hie2022efficient}. We combined measurements for the mAb114 and mAb114UCA antibodies from the original study into a single fitness dataset because the parent sequences shared significant overlap.

\section{Results}
\subsection{Capturing the distribution of observed proteins} 
We first evaluate the capacity of \model to capture the distribution of natural sequences. 
In particular, we focused on its ability to predict unobserved natural sequences, quantifying performance in terms of perplexity on a heldout test set.
We find that larger models yield substantially lower perplexities, consistent with the idea that, despite massive model size, we are far from the overfitting regime (Table \ref{tab:testperformance}).
For a sequence $x=(x_1,x_2,\ldots,x_n)$ of $n$ tokens, the perplexity is calculated as $ppl(x) = \exp\left(-\frac{1}{n} \sum_{i=1}^n \ln p(x_i)\right) = \exp\left(-\frac{1}{n} \sum_{i=1}^n \ln(softmax(logits(x)[i]))[x_i]\right)$ where $logits()$ maps each token $x_i$ to a vector of logit values under the causal language model $p$. We report the average perplexity over the held-out partitions of the datasets.

We caution, however, that these results only reflect the capacity of the model to capture the training distribution $p_0$ from which the data was drawn, not necessarily relevant measures of molecular fitness. 
To be more precise and borrowing notation from \citet{weinstein2022non}, let $p^\infty$ be the stationary distribution of the evolutionary process, such that $\log p^\infty$ is proportional to log fitness $\log f$.
Phylogenetic effects, as well as other imbalances in the dataset, can result in a situation where $p_0 \neq p^\infty$, and so accurate estimation of the training data distribution $p_0$ does not necessarily imply accurate estimation of $p^\infty$ or (consequently) $f$.

\begin{figure}[t]
    \centering
    \includegraphics[width=1.0\textwidth]{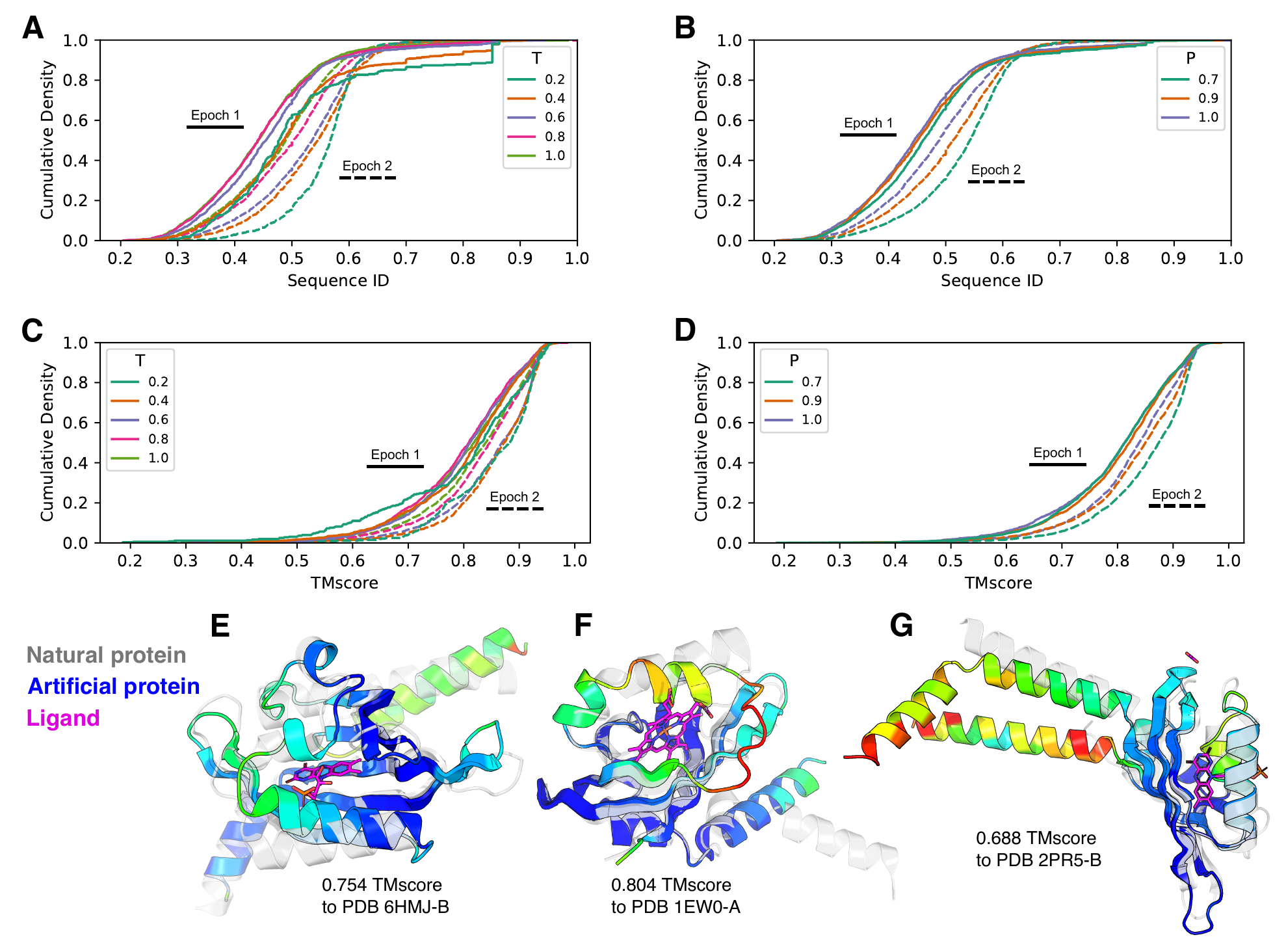}
    \caption{Generating from a language model finetuned on two-layer sandwich architecture proteins. (A-D) Effect of finetuning duration on the sequential and structural similarity of generated proteins to natural proteins. Extended finetuning (two epochs) yields generated sequences more similar to those observed in nature. (A) Higher sampling temperature generates more diverse protein sequences. (B) Higher nucleus-sampling probability produces greater sequence diversity. (C) In general, lower sampling temperature results in sequences adopting structures more similar (higher TMscore) to those found in the PDB. (D) Lower nucleus sampling probability yields generations with reduced structural diversity. (E-G) Comparison of predicted structures for sequences generated by the finetuned language model (colored by pLDDT) and the most structurally similar proteins in the PDB (transparent). Ligands bound by the natural proteins are shown in pink. (E) Generated protein adopting a similar fold to a natural protein binding a flavin mononucleotide ligand. The helical secondary structure of the generated protein matches that of the natural protein near the ligand-binding site, but a shorter loop restricts the space available for binding. (F) Generated protein closely resembling a natural protoporhyrin-binding protein. The structure of the generated protein appears to properly accommodate the ligand, but is predicted with low confidence in the unstructured loop regions near the binding site. (G) Generated protein similar to a natural flavin-mononucleotide-binding protein. The binding site of the generated protein is confidently predicted and reserves appropriate space for the ligand.}
    \label{fig:finetune_generations}
\end{figure}

\subsection{Generation ability}

Given the capacity of the \model family of models for capturing the distribution of observed evolutionary sequences, we next assessed the ability of the models to generate novel sequences. We evaluated sequence generation in three settings: universal protein generation from pretraining, fold-specific generation after finetuning, and antibody generation after domain-specific pretraining.

Prior work has demonstrated that sequences generated by PLMs can adopt a wide variety of folds, often with significant deviation in sequence from observed proteins \citep{madani2020progen, ferruz2022deep}. To assess the generative capacity of \model models, we generated 5,000 sequences with the \model-xlarge model. The three-dimensional structure of each sequences was predicted using AlphaFold2 \citep{jumper2021highly}. For each structure, we identified the most structurally similar natural protein the in the PDB \citep{berman2000protein} using Foldseek \citep{van2022foldseek}. In Figure\ref{fig:pretrain_generations}, we show the relationship between structural similarity to natural proteins (TMscore) and AlphaFold2 prediction confidence (pLDDT). The majority of structures were confidently predicted (median TMscore of 0.89) and had structural homologs in the PDB (median pLDDT of 73.7). However, closer inspection of predicted structures revealed unique several characteristics of the generated sequences. In Figure\ref{fig:pretrain_generations}B, we show a generated sequence adopting a solenoid fold. The closest structural homolog in the PDB is the F-box/LRR-repeat protein 17 (PDB ID 6WCQ-B), a similarly folding solenoid protein. Interestingly, although the inner face of the generated solendoid fold is composed entirely of beta sheets (as in the natural protein), the outer face combines both alpha helices and beta sheets, resulting in a larger central angle (wider curvature). Further, despite adopting similar folds, the sequence identity between the generated and natural proteins is only 11.2\%. For another generated sequence, adopting a multi-domain $\alpha$+$\beta$-fold, no structurally similar proteins were found in the PDB (Figure\ref{fig:pretrain_generations}C). The most similar natural protein was an uncharacterized protein (PDB ID 6OAW-A), with a low TMscore of 0.283 and little sequence overlap (11.1\% identity). In a final case study, we highlight a generated sequence with a predicted structure resembling RTX domain of adenylate cyclase toxin (PDB ID 6SUS-A). Both structures adopt a similar $\beta$-roll fold (TMscore 0.374) and have a  moderate level of sequence similarity (30.3\%). Interestingly, we observe that the generated protein resembles an idealized version of the natural protein, with extended beta sheets and shortened connecting loops. Taken together, these examples illustrate some of the unique properties of sequences generated by \model. While the generated sequences often fold into structures resembling those produced by nature, they frequently do so with significant sequence deviations, and may adopt novel folds in some cases.

Next, we considered generation from a model finetuned on protein sequences adopting a common structural architecture. The \model-large model was finetuned for two epochs on 1M sequences, from Gene3D \citep{lewis2018gene3d} and CATH \citep{sillitoe2021cath}, adopting a two-layer sandwich architecture (CATH 3.30). To understand the effects of extended finetuning, we generated 10,000 sequences using the model parameters after the first and second epoch of finetuning. For all generationed sequence, we calculated the sequence identity against the training dataset using MMseqs2 \citep{steinegger2017mmseqs2}. As expected, we observed higher similarity to observed evolutionary sequences with extended finetuning (Figure\ref{fig:finetune_generations}A-B). Among sequences generated with the same model checkpoints, sampling parameters are strongly correlated with sequence novelty (i.e., higher sampling temperature or nucleus probability yields lower sequence identity). To assess the effect of sampling parameters on structure diversity within the common architecture, we predicted structures for all 20,000 sequences with AlphaFold2 and calculated TMscores against the PDB using Foldseek. A similar trend emerged, with more restrictive sampling parameters typically yielding structures more closely resembling natural proteins (Figure\ref{fig:finetune_generations}C-D). Among the more novel structures, the primary source of diversity is in the ligand-binding regions, while the non-binding regions resemble natural proteins (Figure\ref{fig:finetune_generations}E-G). In two such cases, the ligand-binding region is less confidently predicted by AlphaFold2 and features rearrangements as compared to the closest natural homologs (Figure\ref{fig:finetune_generations}E-F). Interestingly, in both cases the predicted structures present a clear void suitable for a ligand, and even mimic the proximal secondary structures of natural proteins. The lower prediction confidence for these regions could be due to the truncated AlphaFold2 prediction process (one recycle) or the ligand-agnostic nature of the model itself. In another case, the predicted structure of the generated sequence confidently recapitulates the ligand-binding region (Figure\ref{fig:finetune_generations}G). These results demonstrate that the sequences generated by a finetuned model sample diversity at functional regions, while maintaining the common architecture of the training dataset.

\begin{figure}[t]
    \centering
    \includegraphics[width=1.0\textwidth]{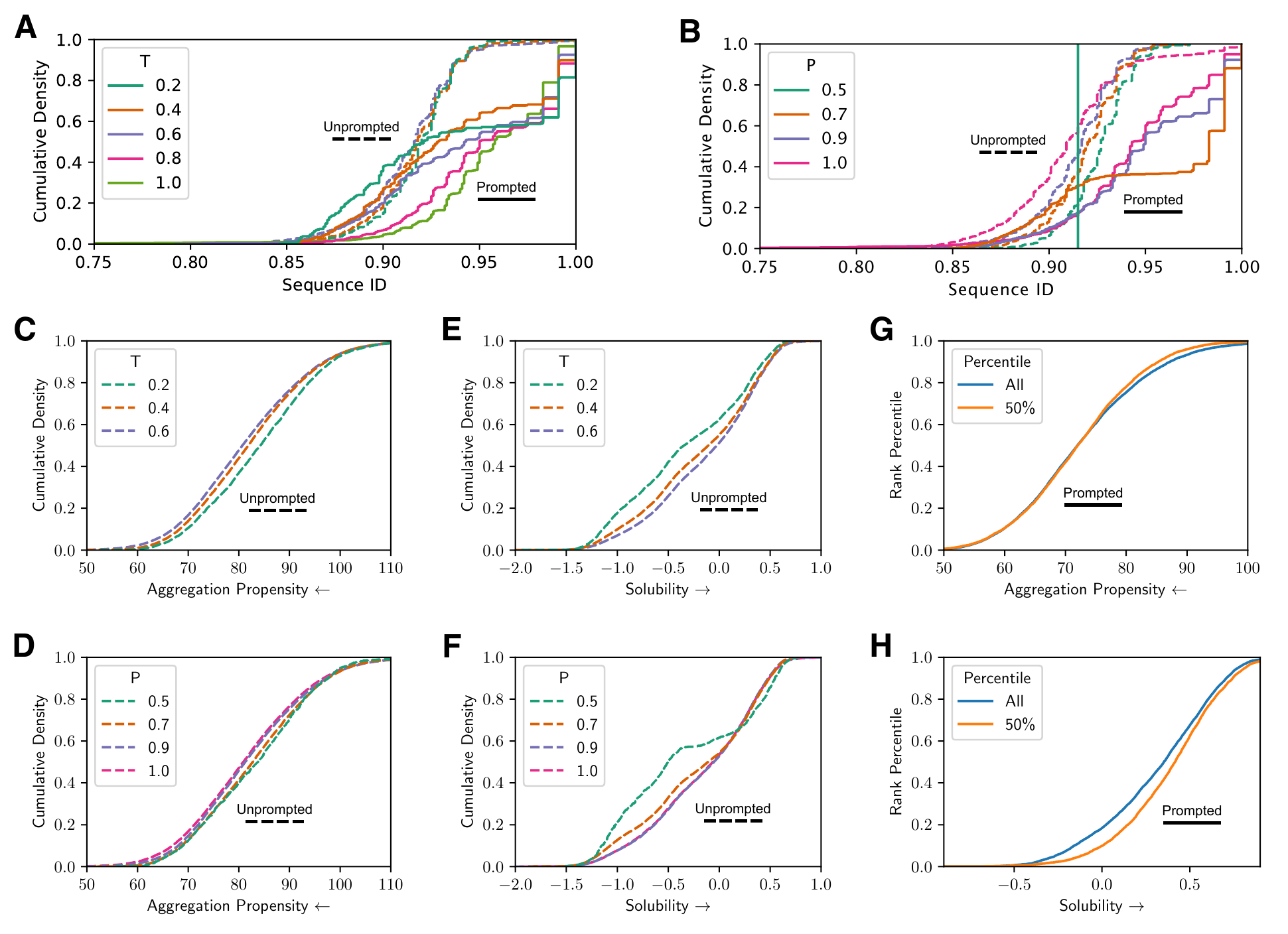}
    \caption{Generating from a pretrained antibody-specific language model. Two strategies were explored for generation of antibody sequences. For unprompted generation, sequences were generated directly by the model without intervention, while for prompted generation the sequence is initialized with three residues (EVQ) commonly observed in human heavy chain antibodies. (A-B) Comparison of sequence identity to the training dataset for unprompted and prompted generations. Prompted generation yields full antibody sequences, resulting in higher sequence identity. (A) Interestingly, higher sampling temperature tends to produce sequences more similar to the training dataset. (B) Lower nucleus sampling probability yields sequences more closely matching the training dataset, as expected. (C-D) Impact of sampling parameters on aggregation propensity of generated sequences. For both temperature sampling (C) and nucleus sampling (D), less restrictive sampling results in lower aggregation propensity for generated sequences. (E-F) Impact of sampling parameters on solubility of sequences. For both temperature sampling (E) and nucleus sampling (F), less restrictive sampling results in more soluble sequences. (G-H) Likelihood ranking of generated antibody sequences with the \model-base language model. (G) Aggregation propensity is not significantly reduced among the top-50\% ranked antibody generations. (H) Solubility is improved by selecting the top 50\% of ranked antibody generations.}
    \label{fig:antibody_generations}
\end{figure}

Generation of antibody sequences is of particular interest for construction of libraries for therapeutic discovery \citep{shin2021protein,shuai2021generative}. However, only relatively small generative models have been trained for this task to date. We investigated the properties of antibody sequences generated by a 764M parameter model pretrained on only natural antibodies. First, we generated 52K non-redundant antibody sequences with the pretrained model. However, experimental limitations of sequencing studies result in over half of antibody sequences in the OAS being truncated at the N-termini by 15 or more residues \citep{olsen2022ablang}. As such, direct generation from the model yields sequences mirroring the training distribution, rather than fully formed antibody sequences. To overcome this bias in the data and produce full-length antibody sequences, we initiated generation with a three-residue motif commonly found at the beginning of human heavy chain sequences (EVQ) \citep{shuai2021generative}. Using this prompting strategy, we generated an additional 470K full-length antibody sequences. In Figure\ref{fig:antibody_generations}A-B, we compare the sequence similarity of unprompted and prompted generations to the training distribution. Notably, the prompted sequences share significantly greater sequence identity with the training distribution, likely due to the inclusion of the highly conserved FW1 region that is frequently absent in the N-terminally-truncated unprompted sequences. Intriguingly, we also observe an inverse relationship between more restrictive sampling parameters (lower temperature, higher nucleus probability) and sequence identity to the training dataset.

Potential antibody therapeutics often require extensive optimization to improve their physical properties. Collectively referred to as developability, these properties include thermal stability, expression, aggregation propensity, and solubility \citep{raybould2019five}. Here, we focused on quantifying the aggregation propensity and solubility of generated sequences according to their SAP scores \citep{chennamsetty2010prediction} and CamSol-intrinsic profiles \citep{sormanni2015camsol}. We found that for both aggregation propensity and solubility, sequences generated with less restrictive parameters display improved developability (Figure\ref{fig:antibody_generations}C-F). Given the effective zero-shot predictive capabilities of PLMs \citep{hesslow2022rita,notin2022tranception}, we also investigated whether a univerally pretrained model could be used to filter generated antibody libraries and improve their developability profiles. In Figure\ref{fig:antibody_generations}G-H, we compare the aggregation propensity and solubility of the full set of generated sequences with the top-50\% as scored by the \model-base model. Among the top-ranked sequences, aggregation propensity improves only marginally, while the solubility of the sequences shows a favorable shift. These results provide meaningful guidance for generation of antibody sequence libraries with PLMs. In practice, generating with less restrictive sampling parameters and filtering with a universal PLM should provide the most developable set of sequences.

\subsection{Zero-shot fitness prediction}
Generative models for protein sequence design should ideally learn a representation that aligns with our desired functional attributes. Experimental techniques in the wet laboratory have allowed for the collection of protein libraries that associate a given sequence to one or many functional scalar values, which describes a \textit{fitness landscape}. We examine how experimentally-measured fitness landscapes correlate with a generative model's likelihood in a zero-shot manner, meaning there is no additional finetuning in a supervised setting with assay-labeled examples or an unsupervised setting with a focused set of homologous sequences.

For a proper comparison to \citet{hesslow2022rita}'s models with a similar architecture to \model yet trained on a different data distribution, we first characterize zero-shot performance on narrow fitness landscapes from \citet{riesselman2018deep} which is comprised mainly of single substitution deep mutational scan experiments. We observe in Table ~\ref{tab:narrow_avg} that our smallest model (\model-small), with an order of magnitude less parameters to RITA-XL, exhibits higher average performance across zero-shot tasks, indicating the importance of pretraining data distributions. In contrast to RITA, the \model training data is a mixture comprised of an identity-reduced set of sequences from Uniref along with sequences from metagenomic sources. Our best \model model outperforms or matches all other baselines spanning a variety of differing modeling strategies-- amplifying the importance of understanding what set of sequences are provided to the model for training.

\begin{table}[t]
\centering
\begin{tabular}{l|cccccc}
\toprule
model       & \begin{tabular}[c]{@{}c@{}}\model-\\ small\end{tabular} & \begin{tabular}[c]{@{}c@{}}\model-\\ base\end{tabular} & \begin{tabular}[c]{@{}c@{}}\model-\\ large\end{tabular}              & \begin{tabular}[c]{@{}c@{}}\model-\\ xlarge\end{tabular}          & \begin{tabular}[c]{@{}c@{}}\model-\\ ensemble\end{tabular} &                                                           \\
average $\rho$ & 0.456                                                    & \textbf{0.505}                                                   & 0.485                                                                 & 0.476                                                              & \textbf{0.518}                                                       &                                                           \\
\toprule
model       & RITA-XL                                                  & EVE                                                     & \begin{tabular}[c]{@{}c@{}}Transception\\ (no retrieval)\end{tabular} & \begin{tabular}[c]{@{}c@{}}Transception\\ (retrieval)\end{tabular} & \begin{tabular}[c]{@{}c@{}}MSA\\ Transformer\end{tabular}   & \begin{tabular}[c]{@{}c@{}}ESM-1v\\ (single)\end{tabular} \\
average $\rho$ & 0.443                                                    & \textbf{0.511}                                                   & 0.447                                                                 & \textbf{0.503}                                                              & 0.476                                                       & 0.475                                                    \\
\bottomrule
\end{tabular}
\vspace{5mm}
\caption{Zero-shot fitness prediction on narrow experimentally-measured fitness landscapes. \model-small outperforms an order of magnitude larger RITA-XL and \model-base is the best performing \model size-- indicating larger model capacity does not always translate to improved predictive performance. \model models outperform or match other baseline methods across a variety of modeling strategies--suggesting the distribution of observed evolutionary sequences provided to the model, along with its inherent biases, likely plays a significant role. The average spearman is reported with data and baselines provided by \citet{hesslow2022rita}}
\label{tab:narrow_avg}
\end{table}

Intriguingly, we find that as model capacity increases, performance at zero-shot fitness prediction (averaged across all datasets in the narrow landscape) peaks at 764M parameters (\model-base) before decreasing with larger and larger models (\model-large and \model-xlarge).
This stands in contrast to model perplexity, which improves systematically with model scale (Table~\ref{tab:testperformance}).
The fact that smaller models, which capture the observed evolutionary data distribution more poorly, can systematically outperform larger models at zero-shot fitness prediction. 
Our results are in line with \citet{weinstein2022non} where the authors show that when $p_0 \neq p^\infty$, fitness estimates from misspecified models can systematically outperform fitness estimates from well-specified models (even in the limit of infinite data), by projecting the data distribution $p_0$ onto a model class closer to $p^\infty$ than $p_0$ itself.
Intuitively, this result says that phylogenetic biases and other distortions in the dataset can be partially corrected for by using a relatively small but well-chosen model, which is capable of describing the key features present in real fitness landscapes but is not capable of exactly matching the data distribution.
Our results provide the first evidence that this effect can hold not only in the context of single protein family datasets but also in the context of large-scale datasets containing evolutionarily diverse proteins, and using large-scale transformer models.

Although bigger models may not translate into better zero-shot fitness performance in general, they may still have advantages in certain cases.
Most of the available fitness assays to which we compare focus on well-studied proteins with large numbers of evolutionarily similar sequences, and measure the fitness/functionality of mutants only one or two mutations away from a wild-type sequence.
Intuitively, regions of sequence space with very low probability under $p_0$ are likely to be especially poorly described with smaller models, and so in these regions both fitness estimation and generation may suffer.
Empirically, we find some suggestive evidence that larger models outperform smaller models at fitness estimation in wider landscapes where sequences are farther from any natural sequence, Table~\ref{tab:wide}.
In particular for the GB1 library, a challenging low-homology protein mutated at positions with non-linear epistasis, our largest models may exhibit emergent behavior \citep{wei2022emergent} in zero-shot identification of the highest fitness variants.

\begin{table}[t]
\centering
\begin{tabular}{l|cccc}
\toprule
dataset {[}metric{]} & \model-small & \model-base    & \model-large   & \model-xlarge  \\
\midrule
AAV {[}AUC{]}        & 0.59        & 0.62          & 0.65          & \textbf{0.68} \\
GFP {[}AUC{]}        & 0.51        & 0.64          & \textbf{0.84} & \textbf{0.84} \\
CM {[}AUC{]}         & 0.68        & \textbf{0.72} & 0.66          & 0.64          \\
GB1 {[}top100avg{]}  & 0.01        & 0.01          & 0.24          & \textbf{0.85} \\
\bottomrule
\end{tabular}
\vspace{5mm}
\caption{Zero-shot fitness prediction on wider experimental landscapes. Larger model capacity may translate to benefits for landscapes involving higher edit distances or low-homology settings. Particularly for GB1 (a low-homology, epistatic landscape), the largest model may demonstrate emergent behavior in finding top ranked sequences.}
\label{tab:wide}
\end{table}

On antibody-specific landscapes, our results again indicate more attention needs to be placed on the distribution of sequences provided to a model during training. We examine the zero-shot fitness prediction of binding ($K_D$) and general properties (expression and melting temperature $T_M$) of antibodies in Table ~\ref{tab:antibody_avg}. Samples from immune repertoire sequencing studies seem like an intuitive choice for learning powerful representations useful for antibody fitness prediction tasks \citep{leem2022deciphering,ruffolo2021deciphering}. However, our \model-OAS model performs poorly as compared to pretrained models trained on universal protein databases. Curiously, the binding prediction performance is non-negligible and may be useful in practical antibody engineering campaigns, even though the corresponding antigen is not provided to the model for likelihood calculation.

\section{Conclusion}
\begin{table}[t]
\centering
\begin{tabular}{l|ccccc}
\toprule
antibody property    & \multicolumn{1}{l}{\begin{tabular}[c]{@{}l@{}}\model-\\ small\end{tabular}} & \multicolumn{1}{l}{\begin{tabular}[c]{@{}l@{}}\model-\\ base\end{tabular}} & \multicolumn{1}{l}{\begin{tabular}[c]{@{}l@{}}\model-\\ large\end{tabular}} & \multicolumn{1}{l}{\begin{tabular}[c]{@{}l@{}}\model-\\ xlarge\end{tabular}} & \multicolumn{1}{l}{\begin{tabular}[c]{@{}l@{}}\model-\\ OAS\end{tabular}} \\
\midrule
binding {[}avg $\rho${]} & \textbf{0.44} & 0.41        & 0.42         & 0.40           & 0.37       \\
general {[}avg $\rho${]} & 0.61          & 0.73        & 0.73         & \textbf{0.74} & 0.66     \\
\bottomrule
\end{tabular}
\vspace{5mm}
\caption{Zero-shot fitness prediction on antibody-specific landscapes. Using redundancy-reduced proteins from immune repertoire sequencing studies, OAS \cite{olsen2022observed}, does not lead to better fitness prediction for antibodies. In particular, we examine antibody fitness predictive performance for binding $K_D$ values and general protein properties including expression quality and $T_M$ melting temperatures. The models trained on universal protein databases are better at predicting general properties as compared to binding affinity. Surprisingly, the binding prediction performance is considerably high considering the associated antigen is not provided to the model.}
\label{tab:antibody_avg}
\end{table}

Protein language models may enable advances in protein engineering and design to solve critical problems for human health and the environment. However, there are many open questions for the field that remain. In our study, the results suggest we can continue to scale model size (>6B parameters) and see appropriate improvements in fitting the distribution of natural sequences. Large protein language models can generate libraries of viable sequences that expand the sequence and structural space of natural proteins. The test-max50 and wide fitness landscape results suggest that scale may particularly show advantages for out-of-distribution, difficult, or tail-end distribution problems. However, our other zero-shot fitness prediction results indicate we need better alignment with respect to data distribution and desired functional predictive ability. Simply using raw sequences as they are collected or reducing redundancy through sequence alignment clustering may not be enough. Worth noting, fitness as defined as an average spearman across the multiple experimental wet lab datasets in this study comes with its own set of biases and may not be the only reliable factor for evaluation of models for protein engineering. We refer the reader to \citep{dallago2021flip,yang2022convolutions} for further discussion. Lastly, we provide our suite of \model models available to enable AI-driven protein engineering research at \href{https://github.com/salesforce/progen}{https://github.com/salesforce/progen}.

\paragraph{Ethics Statement:} Predicting the fitness of a protein sequence and capturing the distribution of natural proteins for generative purposes could be a powerful tool for protein design. If our technique or a future iteration thereof is adopted broadly, care should be taken in terms of the end use-cases of these designed samples and downstream effects to ensure safe, non-nefarious, and ethical applications. For projects in any domain, active oversight during project initiation, experimental optimization, and deployment phases should be put in place to ensure safe usage and limitation of unintended harmful effects.



\bibliography{arxiv_2021}

\begin{thebibliography}{67}
\providecommand{\natexlab}[1]{#1}
\providecommand{\url}[1]{\texttt{#1}}
\expandafter\ifx\csname urlstyle\endcsname\relax
  \providecommand{\doi}[1]{doi: #1}\else
  \providecommand{\doi}{doi: \begingroup \urlstyle{rm}\Url}\fi

\bibitem[Arnold(1998)]{arnold1998design}
Frances~H Arnold.
\newblock Design by directed evolution.
\newblock \emph{Accounts of chemical research}, 31\penalty0 (3):\penalty0
  125--131, 1998.

\bibitem[Bahdanau et~al.(2014)Bahdanau, Cho, and Bengio]{bahdanau2014neural}
Dzmitry Bahdanau, Kyunghyun Cho, and Yoshua Bengio.
\newblock Neural machine translation by jointly learning to align and
  translate.
\newblock \emph{arXiv preprint arXiv:1409.0473}, 2014.

\bibitem[Berman et~al.(2000)Berman, Westbrook, Feng, Gilliland, Bhat, Weissig,
  Shindyalov, and Bourne]{berman2000protein}
Helen~M Berman, John Westbrook, Zukang Feng, Gary Gilliland, Talapady~N Bhat,
  Helge Weissig, Ilya~N Shindyalov, and Philip~E Bourne.
\newblock The protein data bank.
\newblock \emph{Nucleic acids research}, 28\penalty0 (1):\penalty0 235--242,
  2000.

\bibitem[Borgeaud et~al.(2021)Borgeaud, Mensch, Hoffmann, Cai, Rutherford,
  Millican, Driessche, Lespiau, Damoc, Clark, et~al.]{borgeaud2021improving}
Sebastian Borgeaud, Arthur Mensch, Jordan Hoffmann, Trevor Cai, Eliza
  Rutherford, Katie Millican, George van~den Driessche, Jean-Baptiste Lespiau,
  Bogdan Damoc, Aidan Clark, et~al.
\newblock Improving language models by retrieving from trillions of tokens.
\newblock \emph{arXiv preprint arXiv:2112.04426}, 2021.

\bibitem[Bradbury et~al.(2018)Bradbury, Frostig, Hawkins, Johnson, Leary,
  Maclaurin, Necula, Paszke, Vander{P}las, Wanderman-{M}ilne, and
  Zhang]{jax2018github}
James Bradbury, Roy Frostig, Peter Hawkins, Matthew~James Johnson, Chris Leary,
  Dougal Maclaurin, George Necula, Adam Paszke, Jake Vander{P}las, Skye
  Wanderman-{M}ilne, and Qiao Zhang.
\newblock {JAX}: composable transformations of {P}ython+{N}um{P}y programs,
  2018.
\newblock URL \url{http://github.com/google/jax}.

\bibitem[Brandes et~al.(2022)Brandes, Ofer, Peleg, Rappoport, and
  Linial]{brandes2022proteinbert}
Nadav Brandes, Dan Ofer, Yam Peleg, Nadav Rappoport, and Michal Linial.
\newblock Proteinbert: A universal deep-learning model of protein sequence and
  function.
\newblock \emph{Bioinformatics}, 38\penalty0 (8):\penalty0 2102--2110, 2022.

\bibitem[Brown et~al.(2020)Brown, Mann, Ryder, Subbiah, Kaplan, Dhariwal,
  Neelakantan, Shyam, Sastry, Askell, et~al.]{brown2020language}
Tom Brown, Benjamin Mann, Nick Ryder, Melanie Subbiah, Jared~D Kaplan, Prafulla
  Dhariwal, Arvind Neelakantan, Pranav Shyam, Girish Sastry, Amanda Askell,
  et~al.
\newblock Language models are few-shot learners.
\newblock \emph{Advances in neural information processing systems},
  33:\penalty0 1877--1901, 2020.

\bibitem[Chaudhury et~al.(2010)Chaudhury, Lyskov, and
  Gray]{chaudhury2010pyrosetta}
Sidhartha Chaudhury, Sergey Lyskov, and Jeffrey~J Gray.
\newblock Pyrosetta: a script-based interface for implementing molecular
  modeling algorithms using rosetta.
\newblock \emph{Bioinformatics}, 26\penalty0 (5):\penalty0 689--691, 2010.

\bibitem[Chennamsetty et~al.(2010)Chennamsetty, Voynov, Kayser, Helk, and
  Trout]{chennamsetty2010prediction}
Naresh Chennamsetty, Vladimir Voynov, Veysel Kayser, Bernhard Helk, and
  Bernhardt~L Trout.
\newblock Prediction of aggregation prone regions of therapeutic proteins.
\newblock \emph{The Journal of Physical Chemistry B}, 114\penalty0
  (19):\penalty0 6614--6624, 2010.

\bibitem[Dallago et~al.(2021)Dallago, Mou, Johnston, Wittmann, Bhattacharya,
  Goldman, Madani, and Yang]{dallago2021flip}
Christian Dallago, Jody Mou, Kadina~E Johnston, Bruce~J Wittmann, Nicholas
  Bhattacharya, Samuel Goldman, Ali Madani, and Kevin~K Yang.
\newblock Flip: Benchmark tasks in fitness landscape inference for proteins.
\newblock \emph{bioRxiv}, 2021.

\bibitem[Dauparas et~al.(2022)Dauparas, Anishchenko, Bennett, Bai, Ragotte,
  Milles, Wicky, Courbet, de~Haas, Bethel, et~al.]{dauparas2022robust}
J~Dauparas, I~Anishchenko, N~Bennett, H~Bai, RJ~Ragotte, LF~Milles, BIM Wicky,
  A~Courbet, RJ~de~Haas, N~Bethel, et~al.
\newblock Robust deep learning based protein sequence design using proteinmpnn.
\newblock \emph{bioRxiv}, 2022.

\bibitem[Elnaggar et~al.(2020)Elnaggar, Heinzinger, Dallago, Rihawi, Wang,
  Jones, Gibbs, Feher, Angerer, Steinegger, et~al.]{elnaggar2020prottrans}
Ahmed Elnaggar, Michael Heinzinger, Christian Dallago, Ghalia Rihawi, Yu~Wang,
  Llion Jones, Tom Gibbs, Tamas Feher, Christoph Angerer, Martin Steinegger,
  et~al.
\newblock Prottrans: towards cracking the language of life's code through
  self-supervised deep learning and high performance computing.
\newblock \emph{arXiv preprint arXiv:2007.06225}, 2020.

\bibitem[Ferruz et~al.(2022)Ferruz, Schmidt, and H{\"o}cker]{ferruz2022deep}
Noelia Ferruz, Steffen Schmidt, and Birte H{\"o}cker.
\newblock A deep unsupervised language model for protein design.
\newblock \emph{bioRxiv}, 2022.

\bibitem[Fowler \& Fields(2014)Fowler and Fields]{fowler2014deep}
Douglas~M Fowler and Stanley Fields.
\newblock Deep mutational scanning: a new style of protein science.
\newblock \emph{Nature methods}, 11\penalty0 (8):\penalty0 801--807, 2014.

\bibitem[Frazer et~al.(2021)Frazer, Notin, Dias, Gomez, Min, Brock, Gal, and
  Marks]{frazer2021disease}
Jonathan Frazer, Pascal Notin, Mafalda Dias, Aidan Gomez, Joseph~K Min, Kelly
  Brock, Yarin Gal, and Debora~S Marks.
\newblock Disease variant prediction with deep generative models of
  evolutionary data.
\newblock \emph{Nature}, 599\penalty0 (7883):\penalty0 91--95, 2021.

\bibitem[Gumulya et~al.(2018)Gumulya, Baek, Wun, Thomson, Harris, Hunter,
  Behrendorff, Kulig, Zheng, Wu, et~al.]{gumulya2018engineering}
Yosephin Gumulya, Jong-Min Baek, Shun-Jie Wun, Raine~ES Thomson, Kurt~L Harris,
  Dominic~JB Hunter, James~BYH Behrendorff, Justyna Kulig, Shan Zheng, Xueming
  Wu, et~al.
\newblock Engineering highly functional thermostable proteins using ancestral
  sequence reconstruction.
\newblock \emph{Nature Catalysis}, 1\penalty0 (11):\penalty0 878--888, 2018.

\bibitem[Hesslow et~al.(2022)Hesslow, Zanichelli, Notin, Poli, and
  Marks]{hesslow2022rita}
Daniel Hesslow, Niccol{\'o} Zanichelli, Pascal Notin, Iacopo Poli, and Debora
  Marks.
\newblock Rita: a study on scaling up generative protein sequence models.
\newblock \emph{arXiv preprint arXiv:2205.05789}, 2022.

\bibitem[Hie et~al.(2022)Hie, Xu, Shanker, Bruun, Weidenbacher, Tang, and
  Kim]{hie2022efficient}
Brian~L Hie, Duo Xu, Varun~R Shanker, Theodora~UJ Bruun, Payton~A Weidenbacher,
  Shaogeng Tang, and Peter~S Kim.
\newblock Efficient evolution of human antibodies from general protein language
  models and sequence information alone.
\newblock \emph{bioRxiv}, 2022.

\bibitem[Hoffmann et~al.(2022)Hoffmann, Borgeaud, Mensch, Buchatskaya, Cai,
  Rutherford, Casas, Hendricks, Welbl, Clark, et~al.]{hoffmann2022training}
Jordan Hoffmann, Sebastian Borgeaud, Arthur Mensch, Elena Buchatskaya, Trevor
  Cai, Eliza Rutherford, Diego de~Las Casas, Lisa~Anne Hendricks, Johannes
  Welbl, Aidan Clark, et~al.
\newblock Training compute-optimal large language models.
\newblock \emph{arXiv preprint arXiv:2203.15556}, 2022.

\bibitem[Hopf et~al.(2017)Hopf, Ingraham, Poelwijk, Sch{\"a}rfe, Springer,
  Sander, and Marks]{hopf2017mutation}
Thomas~A Hopf, John~B Ingraham, Frank~J Poelwijk, Charlotta~PI Sch{\"a}rfe,
  Michael Springer, Chris Sander, and Debora~S Marks.
\newblock Mutation effects predicted from sequence co-variation.
\newblock \emph{Nature biotechnology}, 35\penalty0 (2):\penalty0 128--135,
  2017.

\bibitem[Hsu et~al.(2022)Hsu, Verkuil, Liu, Lin, Hie, Sercu, Lerer, and
  Rives]{hsu2022learning}
Chloe Hsu, Robert Verkuil, Jason Liu, Zeming Lin, Brian Hie, Tom Sercu, Adam
  Lerer, and Alexander Rives.
\newblock Learning inverse folding from millions of predicted structures.
\newblock \emph{bioRxiv}, 2022.

\bibitem[Ingraham et~al.(2019)Ingraham, Garg, Barzilay, and
  Jaakkola]{ingraham2019generative}
John Ingraham, Vikas Garg, Regina Barzilay, and Tommi Jaakkola.
\newblock Generative models for graph-based protein design.
\newblock \emph{Advances in neural information processing systems}, 32, 2019.

\bibitem[Jing et~al.(2020)Jing, Eismann, Suriana, Townshend, and
  Dror]{jing2020learning}
Bowen Jing, Stephan Eismann, Patricia Suriana, Raphael~JL Townshend, and Ron
  Dror.
\newblock Learning from protein structure with geometric vector perceptrons.
\newblock \emph{arXiv preprint arXiv:2009.01411}, 2020.

\bibitem[Jumper et~al.(2021)Jumper, Evans, Pritzel, Green, Figurnov,
  Ronneberger, Tunyasuvunakool, Bates, {\v{Z}}{\'\i}dek, Potapenko,
  et~al.]{jumper2021highly}
John Jumper, Richard Evans, Alexander Pritzel, Tim Green, Michael Figurnov,
  Olaf Ronneberger, Kathryn Tunyasuvunakool, Russ Bates, Augustin
  {\v{Z}}{\'\i}dek, Anna Potapenko, et~al.
\newblock Highly accurate protein structure prediction with alphafold.
\newblock \emph{Nature}, 596\penalty0 (7873):\penalty0 583--589, 2021.

\bibitem[Kaplan et~al.(2020)Kaplan, McCandlish, Henighan, Brown, Chess, Child,
  Gray, Radford, Wu, and Amodei]{kaplan2020scaling}
Jared Kaplan, Sam McCandlish, Tom Henighan, Tom~B Brown, Benjamin Chess, Rewon
  Child, Scott Gray, Alec Radford, Jeffrey Wu, and Dario Amodei.
\newblock Scaling laws for neural language models.
\newblock \emph{arXiv preprint arXiv:2001.08361}, 2020.

\bibitem[Kingma \& Ba(2015)Kingma and Ba]{kingma2014adam}
Diederik~P. Kingma and Jimmy Ba.
\newblock Adam: A method for stochastic optimization.
\newblock In \emph{ICLR (Poster)}, 2015.
\newblock URL \url{http://arxiv.org/abs/1412.6980}.

\bibitem[Koenig et~al.(2017)Koenig, Lee, Walters, Janakiraman, Stinson,
  Patapoff, and Fuh]{koenig2017mutational}
Patrick Koenig, Chingwei~V Lee, Benjamin~T Walters, Vasantharajan Janakiraman,
  Jeremy Stinson, Thomas~W Patapoff, and Germaine Fuh.
\newblock Mutational landscape of antibody variable domains reveals a switch
  modulating the interdomain conformational dynamics and antigen binding.
\newblock \emph{Proceedings of the National Academy of Sciences}, 114\penalty0
  (4):\penalty0 E486--E495, 2017.

\bibitem[Leaver-Fay et~al.(2011)Leaver-Fay, Tyka, Lewis, Lange, Thompson,
  Jacak, Kaufman, Renfrew, Smith, Sheffler, et~al.]{leaver2011rosetta3}
Andrew Leaver-Fay, Michael Tyka, Steven~M Lewis, Oliver~F Lange, James
  Thompson, Ron Jacak, Kristian~W Kaufman, P~Douglas Renfrew, Colin~A Smith,
  Will Sheffler, et~al.
\newblock Rosetta3: an object-oriented software suite for the simulation and
  design of macromolecules.
\newblock In \emph{Methods in enzymology}, volume 487, pp.\  545--574.
  Elsevier, 2011.

\bibitem[Leem et~al.(2022)Leem, Mitchell, Farmery, Barton, and
  Galson]{leem2022deciphering}
Jinwoo Leem, Laura~S Mitchell, James~HR Farmery, Justin Barton, and Jacob~D
  Galson.
\newblock Deciphering the language of antibodies using self-supervised
  learning.
\newblock \emph{Patterns}, pp.\  100513, 2022.

\bibitem[Lewis et~al.(2018)Lewis, Sillitoe, Dawson, Lam, Clarke, Lee, Orengo,
  and Lees]{lewis2018gene3d}
Tony~E Lewis, Ian Sillitoe, Natalie Dawson, Su~Datt Lam, Tristan Clarke, David
  Lee, Christine Orengo, and Jonathan Lees.
\newblock Gene3d: extensive prediction of globular domains in proteins.
\newblock \emph{Nucleic acids research}, 46\penalty0 (D1):\penalty0 D435--D439,
  2018.

\bibitem[Lu et~al.(2021)Lu, Grover, Abbeel, and Mordatch]{lu2021pretrained}
Kevin Lu, Aditya Grover, Pieter Abbeel, and Igor Mordatch.
\newblock Pretrained transformers as universal computation engines.
\newblock \emph{arXiv preprint arXiv:2103.05247}, 2021.

\bibitem[Madani et~al.(2020)Madani, McCann, Naik, Keskar, Anand, Eguchi, Huang,
  and Socher]{madani2020progen}
Ali Madani, Bryan McCann, Nikhil Naik, Nitish~Shirish Keskar, Namrata Anand,
  Raphael~R Eguchi, Po-Ssu Huang, and Richard Socher.
\newblock Progen: Language modeling for protein generation.
\newblock \emph{arXiv preprint arXiv:2004.03497}, 2020.

\bibitem[Madani et~al.(2021)Madani, Krause, Greene, Subramanian, Mohr, Holton,
  Olmos, Xiong, Sun, Socher, et~al.]{madani2021deep}
Ali Madani, Ben Krause, Eric~R Greene, Subu Subramanian, Benjamin~P Mohr,
  James~M Holton, Jose~Luis Olmos, Caiming Xiong, Zachary~Z Sun, Richard
  Socher, et~al.
\newblock Deep neural language modeling enables functional protein generation
  across families.
\newblock \emph{bioRxiv}, 2021.

\bibitem[Meier et~al.(2021)Meier, Rao, Verkuil, Liu, Sercu, and
  Rives]{meier2021language}
Joshua Meier, Roshan Rao, Robert Verkuil, Jason Liu, Tom Sercu, and Alex Rives.
\newblock Language models enable zero-shot prediction of the effects of
  mutations on protein function.
\newblock \emph{Advances in Neural Information Processing Systems},
  34:\penalty0 29287--29303, 2021.

\bibitem[Nijkamp et~al.(2022)Nijkamp, Pang, Hayashi, Tu, Wang, Zhou, Savarese,
  and Xiong]{nijkamp2022conversational}
Erik Nijkamp, Bo~Pang, Hiroaki Hayashi, Lifu Tu, Huan Wang, Yingbo Zhou, Silvio
  Savarese, and Caiming Xiong.
\newblock A conversational paradigm for program synthesis.
\newblock \emph{arXiv preprint arXiv:2203.13474}, 2022.

\bibitem[Notin et~al.(2022)Notin, Dias, Frazer, Marchena-Hurtado, Gomez, Marks,
  and Gal]{notin2022tranception}
Pascal Notin, Mafalda Dias, Jonathan Frazer, Javier Marchena-Hurtado, Aidan
  Gomez, Debora~S Marks, and Yarin Gal.
\newblock Tranception: protein fitness prediction with autoregressive
  transformers and inference-time retrieval.
\newblock \emph{arXiv preprint arXiv:2205.13760}, 2022.

\bibitem[Olsen et~al.(2022{\natexlab{a}})Olsen, Boyles, and
  Deane]{olsen2022observed}
Tobias~H Olsen, Fergus Boyles, and Charlotte~M Deane.
\newblock Observed antibody space: A diverse database of cleaned, annotated,
  and translated unpaired and paired antibody sequences.
\newblock \emph{Protein Science}, 31\penalty0 (1):\penalty0 141--146,
  2022{\natexlab{a}}.

\bibitem[Olsen et~al.(2022{\natexlab{b}})Olsen, Moal, and
  Deane]{olsen2022ablang}
Tobias~H Olsen, Iain~H Moal, and Charlotte~M Deane.
\newblock Ablang: An antibody language model for completing antibody sequences.
\newblock \emph{bioRxiv}, 2022{\natexlab{b}}.

\bibitem[Pascanu et~al.(2013)Pascanu, Mikolov, and
  Bengio]{pascanu2013difficulty}
Razvan Pascanu, Tomas Mikolov, and Yoshua Bengio.
\newblock On the difficulty of training recurrent neural networks.
\newblock In \emph{International conference on machine learning}, pp.\
  1310--1318. PMLR, 2013.

\bibitem[Rao et~al.(2019)Rao, Bhattacharya, Thomas, Duan, Chen, Canny, Abbeel,
  and Song]{rao2019evaluating}
Roshan Rao, Nicholas Bhattacharya, Neil Thomas, Yan Duan, Peter Chen, John
  Canny, Pieter Abbeel, and Yun Song.
\newblock Evaluating protein transfer learning with tape.
\newblock \emph{Advances in neural information processing systems}, 32, 2019.

\bibitem[Rao et~al.(2021)Rao, Liu, Verkuil, Meier, Canny, Abbeel, Sercu, and
  Rives]{rao2021msa}
Roshan~M Rao, Jason Liu, Robert Verkuil, Joshua Meier, John Canny, Pieter
  Abbeel, Tom Sercu, and Alexander Rives.
\newblock Msa transformer.
\newblock In \emph{International Conference on Machine Learning}, pp.\
  8844--8856. PMLR, 2021.

\bibitem[Raybould et~al.(2019)Raybould, Marks, Krawczyk, Taddese, Nowak, Lewis,
  Bujotzek, Shi, and Deane]{raybould2019five}
Matthew~IJ Raybould, Claire Marks, Konrad Krawczyk, Bruck Taddese, Jaroslaw
  Nowak, Alan~P Lewis, Alexander Bujotzek, Jiye Shi, and Charlotte~M Deane.
\newblock Five computational developability guidelines for therapeutic antibody
  profiling.
\newblock \emph{Proceedings of the National Academy of Sciences}, 116\penalty0
  (10):\penalty0 4025--4030, 2019.

\bibitem[Repecka et~al.(2021)Repecka, Jauniskis, Karpus, Rembeza, Rokaitis,
  Zrimec, Poviloniene, Laurynenas, Viknander, Abuajwa,
  et~al.]{repecka2021expanding}
Donatas Repecka, Vykintas Jauniskis, Laurynas Karpus, Elzbieta Rembeza,
  Irmantas Rokaitis, Jan Zrimec, Simona Poviloniene, Audrius Laurynenas, Sandra
  Viknander, Wissam Abuajwa, et~al.
\newblock Expanding functional protein sequence spaces using generative
  adversarial networks.
\newblock \emph{Nature Machine Intelligence}, 3\penalty0 (4):\penalty0
  324--333, 2021.

\bibitem[Riesselman et~al.(2018)Riesselman, Ingraham, and
  Marks]{riesselman2018deep}
Adam~J Riesselman, John~B Ingraham, and Debora~S Marks.
\newblock Deep generative models of genetic variation capture the effects of
  mutations.
\newblock \emph{Nature methods}, 15\penalty0 (10):\penalty0 816--822, 2018.

\bibitem[Rives et~al.(2021)Rives, Meier, Sercu, Goyal, Lin, Liu, Guo, Ott,
  Zitnick, Ma, et~al.]{rives2021biological}
Alexander Rives, Joshua Meier, Tom Sercu, Siddharth Goyal, Zeming Lin, Jason
  Liu, Demi Guo, Myle Ott, C~Lawrence Zitnick, Jerry Ma, et~al.
\newblock Biological structure and function emerge from scaling unsupervised
  learning to 250 million protein sequences.
\newblock \emph{Proceedings of the National Academy of Sciences}, 118\penalty0
  (15):\penalty0 e2016239118, 2021.

\bibitem[Ruffolo et~al.(2021)Ruffolo, Gray, and Sulam]{ruffolo2021deciphering}
Jeffrey~A Ruffolo, Jeffrey~J Gray, and Jeremias Sulam.
\newblock Deciphering antibody affinity maturation with language models and
  weakly supervised learning.
\newblock \emph{arXiv preprint arXiv:2112.07782}, 2021.

\bibitem[Ruffolo et~al.(2022)Ruffolo, Chu, Mahajan, and Gray]{ruffolo2022fast}
Jeffrey~A Ruffolo, Lee-Shin Chu, Sai~Pooja Mahajan, and Jeffrey~J Gray.
\newblock Fast, accurate antibody structure prediction from deep learning on
  massive set of natural antibodies.
\newblock \emph{bioRxiv}, 2022.

\bibitem[Russ et~al.(2020)Russ, Figliuzzi, Stocker, Barrat-Charlaix, Socolich,
  Kast, Hilvert, Monasson, Cocco, Weigt, et~al.]{russ2020evolution}
William~P Russ, Matteo Figliuzzi, Christian Stocker, Pierre Barrat-Charlaix,
  Michael Socolich, Peter Kast, Donald Hilvert, Remi Monasson, Simona Cocco,
  Martin Weigt, et~al.
\newblock An evolution-based model for designing chorismate mutase enzymes.
\newblock \emph{Science}, 369\penalty0 (6502):\penalty0 440--445, 2020.

\bibitem[Saharia et~al.(2022)Saharia, Chan, Saxena, Li, Whang, Denton,
  Ghasemipour, Ayan, Mahdavi, Lopes, et~al.]{saharia2022photorealistic}
Chitwan Saharia, William Chan, Saurabh Saxena, Lala Li, Jay Whang, Emily
  Denton, Seyed Kamyar~Seyed Ghasemipour, Burcu~Karagol Ayan, S~Sara Mahdavi,
  Rapha~Gontijo Lopes, et~al.
\newblock Photorealistic text-to-image diffusion models with deep language
  understanding.
\newblock \emph{arXiv preprint arXiv:2205.11487}, 2022.

\bibitem[Shin et~al.(2021)Shin, Riesselman, Kollasch, McMahon, Simon, Sander,
  Manglik, Kruse, and Marks]{shin2021protein}
Jung-Eun Shin, Adam~J Riesselman, Aaron~W Kollasch, Conor McMahon, Elana Simon,
  Chris Sander, Aashish Manglik, Andrew~C Kruse, and Debora~S Marks.
\newblock Protein design and variant prediction using autoregressive generative
  models.
\newblock \emph{Nature communications}, 12\penalty0 (1):\penalty0 1--11, 2021.

\bibitem[Shoeybi et~al.(2019)Shoeybi, Patwary, Puri, LeGresley, Casper, and
  Catanzaro]{shoeybi2019megatron}
Mohammad Shoeybi, Mostofa Patwary, Raul Puri, Patrick LeGresley, Jared Casper,
  and Bryan Catanzaro.
\newblock Megatron-lm: Training multi-billion parameter language models using
  model parallelism.
\newblock \emph{arXiv preprint arXiv:1909.08053}, 2019.

\bibitem[Shuai et~al.(2021)Shuai, Ruffolo, and Gray]{shuai2021generative}
Richard~W Shuai, Jeffrey~A Ruffolo, and Jeffrey~J Gray.
\newblock Generative language modeling for antibody design.
\newblock \emph{bioRxiv}, 2021.

\bibitem[Sillitoe et~al.(2021)Sillitoe, Bordin, Dawson, Waman, Ashford,
  Scholes, Pang, Woodridge, Rauer, Sen, et~al.]{sillitoe2021cath}
Ian Sillitoe, Nicola Bordin, Natalie Dawson, Vaishali~P Waman, Paul Ashford,
  Harry~M Scholes, Camilla~SM Pang, Laurel Woodridge, Clemens Rauer, Neeladri
  Sen, et~al.
\newblock Cath: increased structural coverage of functional space.
\newblock \emph{Nucleic acids research}, 49\penalty0 (D1):\penalty0 D266--D273,
  2021.

\bibitem[Sormanni et~al.(2015)Sormanni, Aprile, and
  Vendruscolo]{sormanni2015camsol}
Pietro Sormanni, Francesco~A Aprile, and Michele Vendruscolo.
\newblock The camsol method of rational design of protein mutants with enhanced
  solubility.
\newblock \emph{Journal of molecular biology}, 427\penalty0 (2):\penalty0
  478--490, 2015.

\bibitem[Steinegger \& S{\"o}ding(2017)Steinegger and
  S{\"o}ding]{steinegger2017mmseqs2}
Martin Steinegger and Johannes S{\"o}ding.
\newblock Mmseqs2 enables sensitive protein sequence searching for the analysis
  of massive data sets.
\newblock \emph{Nature biotechnology}, 35\penalty0 (11):\penalty0 1026--1028,
  2017.

\bibitem[Steinegger \& S{\"o}ding(2018)Steinegger and
  S{\"o}ding]{steinegger2018clustering}
Martin Steinegger and Johannes S{\"o}ding.
\newblock Clustering huge protein sequence sets in linear time.
\newblock \emph{Nature communications}, 9\penalty0 (1):\penalty0 1--8, 2018.

\bibitem[Su et~al.(2021)Su, Lu, Pan, Wen, and Liu]{su2021roformer}
Jianlin Su, Yu~Lu, Shengfeng Pan, Bo~Wen, and Yunfeng Liu.
\newblock Roformer: Enhanced transformer with rotary position embedding.
\newblock \emph{arXiv preprint arXiv:2104.09864}, 2021.

\bibitem[Suzek et~al.(2015)Suzek, Wang, Huang, McGarvey, Wu, and
  Consortium]{suzek2015uniref}
Baris~E Suzek, Yuqi Wang, Hongzhan Huang, Peter~B McGarvey, Cathy~H Wu, and
  UniProt Consortium.
\newblock Uniref clusters: a comprehensive and scalable alternative for
  improving sequence similarity searches.
\newblock \emph{Bioinformatics}, 31\penalty0 (6):\penalty0 926--932, 2015.

\bibitem[van Kempen et~al.(2022)van Kempen, Kim, Tumescheit, Mirdita,
  S{\"o}ding, and Steinegger]{van2022foldseek}
Michel van Kempen, Stephanie Kim, Charlotte Tumescheit, Milot Mirdita, Johannes
  S{\"o}ding, and Martin Steinegger.
\newblock Foldseek: fast and accurate protein structure search.
\newblock \emph{bioRxiv}, 2022.

\bibitem[Vaswani et~al.(2017)Vaswani, Shazeer, Parmar, Uszkoreit, Jones, Gomez,
  Kaiser, and Polosukhin]{vaswani2017attention}
Ashish Vaswani, Noam Shazeer, Niki Parmar, Jakob Uszkoreit, Llion Jones,
  Aidan~N Gomez, {\L}ukasz Kaiser, and Illia Polosukhin.
\newblock Attention is all you need.
\newblock \emph{Advances in neural information processing systems}, 30, 2017.

\bibitem[Wang \& Komatsuzaki(2021)Wang and Komatsuzaki]{gpt-j}
Ben Wang and Aran Komatsuzaki.
\newblock {GPT-J-6B: A 6 Billion Parameter Autoregressive Language Model}.
\newblock \url{https://github.com/kingoflolz/mesh-transformer-jax}, May 2021.

\bibitem[Warszawski et~al.(2019)Warszawski, Borenstein~Katz, Lipsh,
  Khmelnitsky, Ben~Nissan, Javitt, Dym, Unger, Knop, Albeck,
  et~al.]{warszawski2019optimizing}
Shira Warszawski, Aliza Borenstein~Katz, Rosalie Lipsh, Lev Khmelnitsky, Gili
  Ben~Nissan, Gabriel Javitt, Orly Dym, Tamar Unger, Orli Knop, Shira Albeck,
  et~al.
\newblock Optimizing antibody affinity and stability by the automated design of
  the variable light-heavy chain interfaces.
\newblock \emph{PLoS computational biology}, 15\penalty0 (8):\penalty0
  e1007207, 2019.

\bibitem[Wei et~al.(2022)Wei, Tay, Bommasani, Raffel, Zoph, Borgeaud, Yogatama,
  Bosma, Zhou, Metzler, et~al.]{wei2022emergent}
Jason Wei, Yi~Tay, Rishi Bommasani, Colin Raffel, Barret Zoph, Sebastian
  Borgeaud, Dani Yogatama, Maarten Bosma, Denny Zhou, Donald Metzler, et~al.
\newblock Emergent abilities of large language models.
\newblock \emph{arXiv preprint arXiv:2206.07682}, 2022.

\bibitem[Weinstein et~al.(2022)Weinstein, Amin, Frazer, and
  Marks]{weinstein2022non}
Eli~N Weinstein, Alan~N Amin, Jonathan Frazer, and Debora~S Marks.
\newblock Non-identifiability and the blessings of misspecification in models
  of molecular fitness and phylogeny.
\newblock \emph{bioRxiv}, 2022.

\bibitem[Wicky et~al.(2022)Wicky, Milles, Courbet, Ragotte, Dauparas, Kinfu,
  Tipps, Kibler, Baek, DiMaio, et~al.]{wicky2022hallucinating}
Basile~IM Wicky, Lukas~F Milles, Alexis Courbet, Robert~J Ragotte, Justas
  Dauparas, Elias Kinfu, Sam Tipps, Ryan~D Kibler, Minkyung Baek, Frank DiMaio,
  et~al.
\newblock Hallucinating protein assemblies.
\newblock \emph{bioRxiv}, 2022.

\bibitem[Yang et~al.(2022)Yang, Lu, and Fusi]{yang2022convolutions}
Kevin~K Yang, Alex~X Lu, and Nicolo~K Fusi.
\newblock Convolutions are competitive with transformers for protein sequence
  pretraining.
\newblock \emph{bioRxiv}, 2022.

\bibitem[Zhang \& Skolnick(2004)Zhang and Skolnick]{zhang2004scoring}
Yang Zhang and Jeffrey Skolnick.
\newblock Scoring function for automated assessment of protein structure
  template quality.
\newblock \emph{Proteins: Structure, Function, and Bioinformatics}, 57\penalty0
  (4):\penalty0 702--710, 2004.

\end{thebibliography}
\bibliographystyle{arxiv_2021}

\newpage

\appendix
\section{Supplementary Material}

\begin{table}[ht]
\centering
\begin{tabular}{l|cccc}
\toprule
\multicolumn{1}{c}{dataset}  & \begin{tabular}[c]{@{}c@{}}\model-\\ base\end{tabular} & tranception & \begin{tabular}[c]{@{}c@{}}tranception \\ (retrieval)\end{tabular} & wavenet     \\
\midrule
A0A1J4YT16\_9PROT\_Davidi\_2020       & \textbf{0.195}                                                   & 0.178                & \textbf{0.191}                                                              & 0.117                \\
B1LPA6\_ECOSM\_Russ\_2020             & \textbf{0.405}                                                   & 0.321                & \textbf{0.415}                                                              & 0.385                \\
BLAT\_ECOLX\_indels                   & \textbf{0.664}                                                   & 0.296                & 0.357                                                                       & 0.546                \\
PTEN\_HUMAN\_Mighell\_2018\_deletions & 0.641                                                            & 0.563                & 0.598                                                                       & \textbf{0.699}       \\
CAPSD\_AAV2S\_Sinai\_2021\_indels     & 0.392                                                            & 0.549                & \textbf{0.586}                                                              & 0.457                \\
HIS7\_YEAST\_Pokusaeva\_2019\_indels  & \textbf{0.702}                                                   & \textbf{0.707}       & \textbf{0.692}                                                              & 0.68                 \\
P53\_HUMAN\_Kotler\_2018\_deletions   & \textbf{0.398}                                                   & \textbf{0.395}       & \textbf{0.401}                                                              & 0.001                \\
\midrule
\textbf{AVERAGE}                               & \textbf{0.485}                                                   & 0.430                & 0.463                                                                       & 0.412 \\
\bottomrule
\end{tabular}
\vspace{4mm}
\caption{Zero-shot fitness prediction on experimental studies evaluating indels collected by \cite{notin2022tranception}. \model outperforms baselines including models with inference-time retrieval.}
\label{tab:indels}
\end{table}

\begin{table}[ht]
\centering
\begin{tabular}{l|ccccc}
\toprule
antibody property    & \multicolumn{1}{l}{\begin{tabular}[c]{@{}l@{}}\model-\\ small\end{tabular}} & \multicolumn{1}{l}{\begin{tabular}[c]{@{}l@{}}\model-\\ base\end{tabular}} & \multicolumn{1}{l}{\begin{tabular}[c]{@{}l@{}}\model-\\ large\end{tabular}} & \multicolumn{1}{l}{\begin{tabular}[c]{@{}l@{}}\model-\\ xlarge\end{tabular}} & \multicolumn{1}{l}{\begin{tabular}[c]{@{}l@{}}\model-\\ OAS\end{tabular}} \\
\midrule
g6\_bind             & \textbf{0.164}                                                              & \textbf{0.153}                                                             & \textbf{0.152}                                                              & 0.135                                                                        & 0.032                                                                     \\
g6\_exp              & \textbf{0.411}                                                              & 0.386                                                                      & 0.368                                                                       & 0.352                                                                        & 0.184                                                                     \\
d44\_bind            & 0.290                                                                       & \textbf{0.426}                                                             & \textbf{0.421}                                                              & 0.366                                                                        & 0.260                                                                     \\
C143\_Kd             & 0.162                                                                       & \textbf{0.197}                                                             & 0.132                                                                       & 0.085                                                                        & 0.006                                                                     \\
C143\_Tm             & \textbf{1.000}                                                              & \textbf{1.000}                                                             & \textbf{1.000}                                                              & \textbf{1.000}                                                               & \textbf{1.000}                                                            \\
MEDI8852UCA\_Kd      & 0.466                                                                       & \textbf{0.913}                                                             & \textbf{0.916}                                                              & \textbf{0.902}                                                               & 0.798                                                                     \\
MEDI8852UCA\_Tm      & 0.314                                                                       & \textbf{0.829}                                                             & \textbf{0.829}                                                              & \textbf{0.829}                                                               & 0.771                                                                     \\
MEDI8852\_Kd         & \textbf{0.504}                                                              & 0.007                                                                      & 0.021                                                                       & 0.043                                                                        & 0.157                                                                     \\
MEDI8852\_Tm         & \textbf{1.000}                                                              & \textbf{1.000}                                                             & \textbf{1.000}                                                              & \textbf{1.000}                                                               & \textbf{1.000}                                                            \\
REGN10987\_Kd        & \textbf{0.520}                                                              & 0.327                                                                      & 0.377                                                                       & 0.462                                                                        & 0.396                                                                     \\
REGN10987\_Tm        & 0.238                                                                       & \textbf{0.810}                                                             & 0.762                                                                       & 0.667                                                                        & 0.667                                                                     \\
S309\_Kd             & \textbf{0.728}                                                              & 0.526                                                                      & 0.586                                                                       & 0.581                                                                        & 0.600                                                                     \\
S309\_Tm             & \textbf{0.830}                                                              & 0.770                                                                      & 0.758                                                                       & 0.794                                                                        & 0.794                                                                     \\
mAb114-mAb114UCA\_Kd & 0.658                                                                       & \textbf{0.769}                                                             & 0.727                                                                       & 0.658                                                                        & 0.738                                                                     \\
mAb114-mAb114UCA\_Tm & 0.500                                                                       & 0.333                                                                      & 0.383                                                                       & \textbf{0.517}                                                               & 0.200                                                                     \\
\midrule
\textbf{AVERAGE}     & 0.519                                                                       & \textbf{0.563}                                                             & \textbf{0.562}                                                              & \textbf{0.559}                                                               & 0.507 \\
\bottomrule
\end{tabular}
\vspace{4mm}
\caption{Full results of zero-shot fitness prediction for the antibody landscapes evaluated in this study.}
\label{tab:antibody_full}
\end{table}

\begin{table}[ht]
\centering
\begin{tabular}{l|cc}
\toprule
\multicolumn{1}{c}{}  & Uniref90+BFD30 & Uniref90+BFD90 \\
\midrule
test set {[}ppl{]}   & 12.7              & \textbf{12.6}     \\
antibody binding {[}rho{]} & \textbf{0.42}     & 0.38              \\
antibody general {[}rho{]} & 0.73              & \textbf{0.76}     \\
narrow landscape avg {[}rho{]} & 0.49              & \textbf{0.50}     \\
AAV {[}auc{]}         & 0.65              & \textbf{0.67}     \\
GFP {[}auc{]}         & \textbf{0.84}     & 0.79              \\
CM {[}auc{]}          & \textbf{0.66}     & 0.65              \\
GB1 {[}top100avg{]}   & 0.24              & \textbf{0.33}    \\
\bottomrule
\end{tabular}
\vspace{4mm}
\caption{Comparing language modeling and zero-shot fitness prediction performance for two 2.7B parameter models trained with differing amounts of proteins from metagenomic sources. BFD30 and BFD90 are clustered at 30\% and 90\% sequence identity respectively. The Uniref90+BFD90 database is majority metagenomic.}
\label{tab:bfd30v90}
\end{table}

\newpage
\begin{landscape}
\begin{table}[ht]
\centering
\begin{tabular}{l|ccccccccccc}
\toprule
\multicolumn{1}{c}{dataset}   & \begin{tabular}[c]{@{}c@{}}\model-\\ small\end{tabular} & \begin{tabular}[c]{@{}c@{}}\model-\\ base\end{tabular} & \begin{tabular}[c]{@{}c@{}}\model-\\ large\end{tabular} & \begin{tabular}[c]{@{}c@{}}\model-\\ xlarge\end{tabular} & \begin{tabular}[c]{@{}c@{}}\model-\\ ensemble\end{tabular} & rita-xl & eve            & \begin{tabular}[c]{@{}c@{}}transcept\end{tabular} & \begin{tabular}[c]{@{}c@{}}transcept\\ (retrieval)\end{tabular} & \begin{tabular}[c]{@{}c@{}}msa \\ transf\end{tabular} & \begin{tabular}[c]{@{}c@{}}esm-1v\\ (single)\end{tabular} \\
\midrule
A0A192B1T2 Haddox 2018  & 0.487                                                    & 0.441                                                   & 0.483                                                    & 0.460                                                     & 0.490                                                       & 0.500   & 0.510          & 0.510                                                                  & 0.509                                                              & 0.510                                                      & 0.490                                                     \\
A0A2Z5U3Z0 9INFA Doud 2016    & 0.486                                                    & 0.573                                                   & 0.555                                                    & 0.563                                                     & 0.584                                                       & 0.540   & 0.534          & 0.539                                                                  & 0.549                                                              & 0.160                                                      & 0.510                                                     \\
AMIE PSEAE Wrenbeck 2017      & 0.584                                                    & 0.558                                                   & 0.580                                                    & 0.549                                                     & 0.593                                                       & 0.560   & 0.558          & 0.501                                                                  & 0.585                                                              & 0.610                                                      & 0.610                                                     \\
B3VI55 LIPST Klesmith 2015    & 0.425                                                    & 0.584                                                   & 0.538                                                    & 0.596                                                     & 0.591                                                       & 0.450   & 0.436          & 0.491                                                                  & 0.468                                                              & 0.530                                                      & 0.480                                                     \\
BLAT ECOLX Deng 2012          & 0.451                                                    & 0.533                                                   & 0.480                                                    & 0.439                                                     & 0.529                                                       & 0.400   & 0.506          & 0.381                                                                  & 0.480                                                              & 0.560                                                      & 0.530                                                     \\
BLAT ECOLX Firnberg 2014      & 0.606                                                    & 0.684                                                   & 0.602                                                    & 0.505                                                     & 0.672                                                       & 0.560   & 0.741          & 0.521                                                                  & 0.678                                                              & 0.740                                                      & 0.680                                                     \\
BLAT ECOLX Jacquier 2013      & 0.564                                                    & 0.636                                                   & 0.574                                                    & 0.478                                                     & 0.647                                                       & 0.560   & 0.742          & 0.539                                                                  & 0.683                                                              & 0.700                                                      & 0.680                                                     \\
BLAT ECOLX Stiffler 2015      & 0.586                                                    & 0.666                                                   & 0.569                                                    & 0.470                                                     & 0.651                                                       & 0.570   & 0.740          & 0.518                                                                  & 0.667                                                              & 0.730                                                      & 0.690                                                     \\
BRCA1 HUMAN Findlay 2018      & 0.245                                                    & 0.441                                                   & 0.425                                                    & 0.455                                                     & 0.460                                                       & 0.070   & 0.322          & 0.538                                                                  & 0.574                                                              & 0.400                                                      & 0.440                                                     \\
CALM1 HUMAN Weile 2017        & 0.292                                                    & 0.331                                                   & 0.349                                                    & 0.354                                                     & 0.356                                                       & 0.270   & 0.244          & 0.306                                                                  & 0.283                                                              & 0.254                                                      & 0.246                                                     \\
DLG4 RAT McLaughlin 2012      & 0.440                                                    & 0.456                                                   & 0.364                                                    & 0.351                                                     & 0.417                                                       & 0.370   & 0.523          & 0.304                                                                  & 0.446                                                              & 0.507                                                      & 0.565                                                     \\
GAL4 YEAST Kitzman 2015       & 0.349                                                    & 0.466                                                   & 0.437                                                    & 0.569                                                     & 0.517                                                       & 0.350   & 0.511          & 0.326                                                                  & 0.526                                                              & 0.583                                                      & 0.441                                                     \\
HSP82 YEAST Mishra 2016       & 0.528                                                    & 0.559                                                   & 0.499                                                    & 0.494                                                     & 0.536                                                       & 0.530   & 0.537          & 0.503                                                                  & 0.530                                                              & 0.482                                                      & 0.568                                                     \\
IF1 ECOLI Kelsic 2016         & 0.485                                                    & 0.513                                                   & 0.477                                                    & 0.492                                                     & 0.522                                                       & 0.420   & 0.525          & 0.548                                                                  & 0.509                                                              & 0.227                                                      & 0.538                                                     \\
KKA2 KLEPN Melnikov 2014      & 0.323                                                    & 0.538                                                   & 0.534                                                    & 0.584                                                     & 0.593                                                       & 0.560   & 0.597          & 0.584                                                                  & 0.584                                                              & 0.576                                                      & 0.614                                                     \\
MK01 HUMAN Brenan 2016        & 0.182                                                    & 0.057                                                   & 0.076                                                    & 0.068                                                     & 0.048                                                       & 0.050   & 0.251          & 0.034                                                                  & 0.139                                                              & 0.153                                                      & 0.183                                                     \\
MTH3 HAEAE Rockah 2015 & 0.479                                                    & 0.711                                                   & 0.684                                                    & 0.710                                                     & 0.717                                                       & 0.680   & 0.710          & 0.673                                                                  & 0.655                                                              & 0.687                                                      & 0.701                                                     \\
P84126 THETH Chan 2017        & 0.501                                                    & 0.573                                                   & 0.560                                                    & 0.657                                                     & 0.635                                                       & 0.550   & 0.567          & 0.533                                                                  & 0.541                                                              & 0.631                                                      & 0.546                                                     \\
PABP YEAST Melamed 2013       & 0.555                                                    & 0.633                                                   & 0.639                                                    & 0.644                                                     & 0.654                                                       & 0.680   & 0.639          & 0.641                                                                  & 0.689                                                              & 0.662                                                      & 0.665                                                     \\
PA I34A1 Wu 2015              & 0.301                                                    & 0.572                                                   & 0.564                                                    & 0.569                                                     & 0.572                                                       & 0.540   & 0.539          & 0.541                                                                  & 0.572                                                              & 0.383                                                      & 0.054                                                     \\
POLG HCVJF Qi 2014            & 0.524                                                    & 0.522                                                   & 0.513                                                    & 0.589                                                     & 0.562                                                       & 0.440   & 0.614          & 0.525                                                                  & 0.577                                                              & 0.600                                                      & 0.605                                                     \\
Q2N0S5 9HIV1 Haddox 2018      & 0.510                                                    & 0.372                                                   & 0.423                                                    & 0.319                                                     & 0.412                                                       & 0.350   & 0.496          & 0.412                                                                  & 0.492                                                              & 0.490                                                      & 0.496                                                     \\
Q59976 STRSQ Romero 2015      & 0.708                                                    & 0.759                                                   & 0.756                                                    & 0.756                                                     & 0.771                                                       & 0.650   & 0.647          & 0.645                                                                  & 0.659                                                              & 0.674                                                      & 0.506                                                     \\
RASH HUMAN Bandaru 2017       & 0.429                                                    & 0.410                                                   & 0.375                                                    & 0.319                                                     & 0.408                                                       & 0.400   & 0.454          & 0.377                                                                  & 0.447                                                              & 0.415                                                      & 0.359                                                     \\
RL401 YEAST Mavor 2016        & 0.482                                                    & 0.454                                                   & 0.478                                                    & 0.460                                                     & 0.486                                                       & 0.430   & 0.364          & 0.331                                                                  & 0.368                                                              & 0.378                                                      & 0.297                                                     \\
RL401 YEAST Roscoe 2013       & 0.552                                                    & 0.547                                                   & 0.549                                                    & 0.539                                                     & 0.566                                                       & 0.490   & 0.421          & 0.392                                                                  & 0.419                                                              & 0.434                                                      & 0.314                                                     \\
RL401 YEAST Roscoe 2014       & 0.426                                                    & 0.399                                                   & 0.379                                                    & 0.364                                                     & 0.408                                                       & 0.390   & 0.401          & 0.343                                                                  & 0.404                                                              & 0.365                                                      & 0.254                                                     \\
SUMO1 HUMAN Weile 2017        & 0.532                                                    & 0.536                                                   & 0.513                                                    & 0.434                                                     & 0.513                                                       & 0.390   & 0.531          & 0.424                                                                  & 0.488                                                              & 0.423                                                      & 0.430                                                     \\
TPK1 HUMAN Weile 2017         & 0.181                                                    & 0.360                                                   & 0.335                                                    & 0.382                                                     & 0.380                                                       & 0.290   & 0.230          & 0.313                                                                  & 0.314                                                              & 0.268                                                      & 0.284                                                     \\
TPMT HUMAN Matreyek 2018      & 0.451                                                    & 0.494                                                   & 0.478                                                    & 0.441                                                     & 0.513                                                       & 0.510   & 0.548          & 0.445                                                                  & 0.522                                                              & 0.508                                                      & 0.540                                                     \\
TRPC SACS2 Chan 2017          & 0.503                                                    & 0.559                                                   & 0.525                                                    & 0.517                                                     & 0.573                                                       & 0.570   & 0.577          & 0.551                                                                  & 0.585                                                              & 0.629                                                      & 0.606                                                     \\
TRPC THEMA Chan 2017          & 0.475                                                    & 0.377                                                   & 0.415                                                    & 0.474                                                     & 0.425                                                       & 0.450   & 0.420          & 0.453                                                                  & 0.436                                                              & 0.474                                                      & 0.472                                                     \\
UBC9 HUMAN Weile 2017         & 0.481                                                    & 0.525                                                   & 0.507                                                    & 0.499                                                     & 0.524                                                       & 0.450   & 0.538          & 0.433                                                                  & 0.485                                                              & 0.503                                                      & 0.479                                                     \\
UBE4B MOUSE Starita 2013      & 0.452                                                    & 0.429                                                   & 0.410                                                    & 0.292                                                     & 0.420                                                       & 0.300   & 0.476          & 0.256                                                                  & 0.388                                                              & 0.347                                                      & 0.462                                                     \\
YAP1 HUMAN Araya 2012         & 0.368                                                    & 0.404                                                   & 0.344                                                    & 0.281                                                     & 0.386                                                       & 0.190   & 0.438          & 0.218                                                                  & 0.359                                                              & 0.071                                                      & 0.281                                                     \\
\midrule
\textbf{AVERAGE}              & 0.456                                                    & \textbf{0.505}                                          & 0.485                                                    & 0.476                                                     & \textbf{0.518}                                              & 0.443   & \textbf{0.511} & 0.447                                                                  & \textbf{0.503}                                                     & 0.476                                                      & 0.475       \\
\bottomrule
\end{tabular}
\vspace{2mm}
\caption{Full results of zero-shot fitness prediction for narrow landscapes.}
\label{tab:narrow_full}
\end{table}
\end{landscape}
\end{document}